\documentclass[conference]{IEEEtran}

\usepackage{times}
\usepackage{multicol}
\usepackage[bookmarks=true]{hyperref}
\usepackage{url}
\usepackage{graphicx}
\usepackage{amsthm}
\usepackage{array}
\usepackage{xcolor}
\usepackage[footnotesize]{caption} %makes captions have different sizes
\usepackage{subcaption}
\usepackage[linesnumbered,ruled,vlined]{algorithm2e}
\usepackage{multirow}
\usepackage{booktabs}
\usepackage{marginnote}
\usepackage[normalem]{ulem}

\usepackage{amsmath,amsfonts,bm}

\def\1{\bm{1}}

\DeclareMathAlphabet{\mathsfit}{\encodingdefault}{\sfdefault}{m}{sl}
\SetMathAlphabet{\mathsfit}{bold}{\encodingdefault}{\sfdefault}{bx}{n}

\newcommand{\E}{\mathbb{E}}

\newcommand{\R}{\mathbb{R}}

\newtheorem{defn}{Definition}

\newtheorem{thm}[defn]{Theorem}

\newcommand{\N}{\mathcal{N}}

\newcommand{\Z}{\mathcal{Z}}

\newcommand{\NIG}{\mathcal{N}\Gamma^{-1}}

\SetKwInput{KwInput}{Initialize}
\SetKwInput{KwOutput}{Output}
\SetKwInOut{Requires}{Requires}
\SetKwInOut{Return}{Return}
\SetKwInOut{Algorithm}{Algorithm}

\newcommand{\reviewremove}[1]{\textnormal{{\color{red}\sout{}}}}
\newcommand{\reviewadd}[1]{\textnormal{{#1}}}

\usepackage[
    style=ieee,
    doi=false,
    isbn=false,
    url=false,
    eprint=false,
    backend=bibtex,
    natbib=true
    ]{biblatex}
    
\bibliography{references}
\IEEEoverridecommandlockouts
\begin{document}

% paper title
\title{You've Got to Feel It To Believe It: Multi-Modal Bayesian Inference  
 for Semantic and Property Prediction}

\author{\authorblockN{Parker Ewen,
Hao Chen,
Yuzhen Chen, 
Anran Li,
Anup Bagali,
Gitesh Gunjal, and
Ram Vasudevan}
\authorblockA{University of Michigan, Ann Arbor, USA}
\thanks{This work is supported by the Ford Motor Company and the National Science Foundation Career Award \#1751093.}
}

\maketitle

\begin{abstract}
Robots must be able to understand their surroundings to perform complex tasks in challenging environments and many of these complex tasks require estimates of physical properties such as friction or weight.
Estimating such properties using learning is challenging due to the large amounts of labelled data required for training and the difficulty of updating these learned models online at run time.
To overcome these challenges, this paper introduces a novel, multi-modal approach for representing semantic predictions and physical property estimates jointly in a probabilistic manner.
By using conjugate pairs, the proposed method enables closed-form Bayesian updates given visual and tactile measurements without requiring additional training data.
The efficacy of the proposed algorithm is demonstrated through several simulation and hardware experiments.
In particular, this paper illustrates that by conditioning semantic classifications on physical properties, the proposed method quantitatively outperforms state-of-the-art semantic classification methods that rely on vision alone.
To further illustrate its utility, the proposed method is used in several applications including to represent affordance-based properties probabilistically and a challenging terrain traversal task using a legged robot.
In the latter task, the proposed method represents the coefficient of friction of the terrain probabilistically, which enables the use of an on-line risk-aware planner that switches the legged robot from a dynamic gait to a static, stable gait when the expected value of the coefficient of friction falls below a given threshold.
Videos of these case studies as well as the open-source C++ and ROS interface can be found at \href{https://roahmlab.github.io/multimodal_mapping/}{https://roahmlab.github.io/multimodal\_mapping/}
\end{abstract}

\IEEEpeerreviewmaketitle

\section{Introduction} \label{sec:intro}
Scene understanding from exteroceptive sensing via images or point clouds enables mobile robots to perform object avoidance, terrain traversal, and a variety of other tasks.
The introduction of high-level and task-dependent semantic labels for scene understanding has recently spurred rapid growth in this area \cite{guo2018review, vaswani2017attention}.
To extract semantic labels from images or point clouds, it is common to employ semantic segmentation neural networks; however, even state-of-the-art methods return inconsistent labels under viewpoint or lighting changes, and are heavily reliant on the data used for training.
Unfortunately, collecting data to train these networks is expensive and out-of-distribution examples may lead to incorrect classifications.

To address the challenge of viewpoint inconsistency, several state-of-the-art methods project image-based semantic classifications onto metric maps \cite{gan2022multitask, mccormac2017semanticfusion}.
This provides a shared frame of reference between images thereby enabling the application of probabilistic methods to fuse semantic data.
While such methods demonstrate improved performance over state-of-the-art image-based segmentation \cite{ewen2022these}, these methods have been restricted to either visual or lidar data.
This is limiting since the properties of a semantic class are often of primary interest rather than the class labels themselves while completing a robotic task.
For instance, the coefficient of friction is more informative than the terrain class for a legged robot.
While recent semantic mapping methods have derived recursive filters for image-based semantic predictions, properties conditioned on these semantic classes are often assumed to be known {\it a priori} and are considered immutable at run time \cite{cai2023evora, ewen2022these, gan2022multitask}.

\begin{figure} [!t]
    \centering
    \includegraphics[width=\columnwidth]{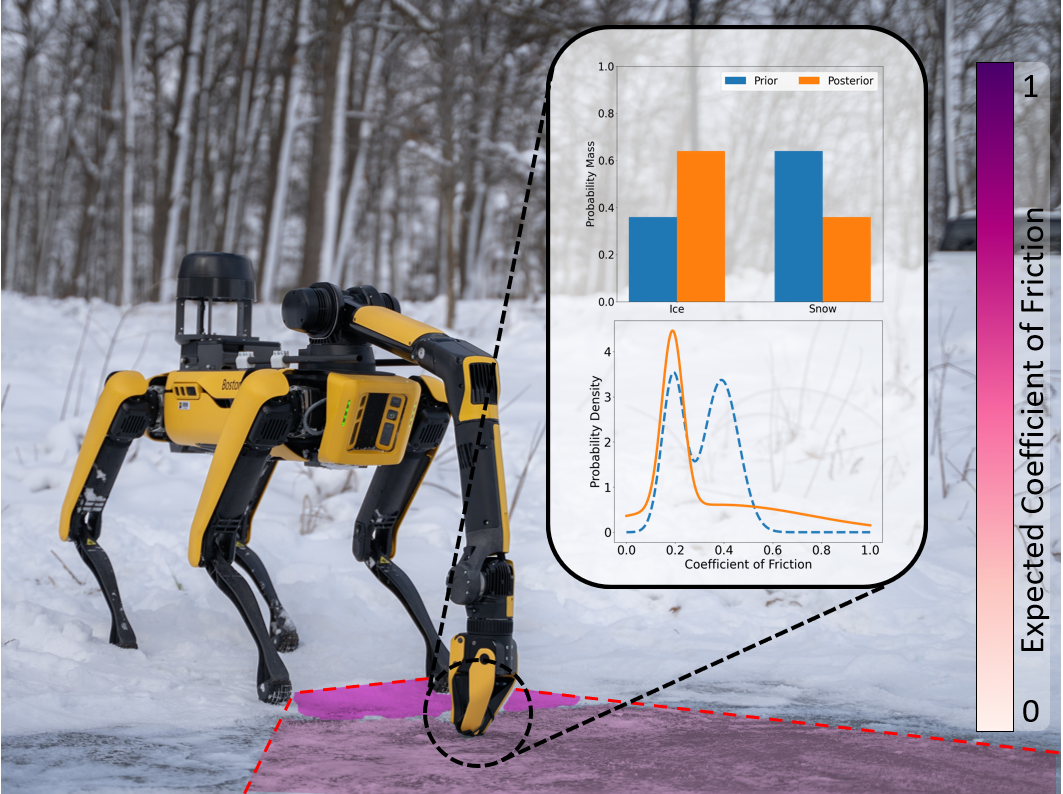}
    \caption{The method proposed in this paper jointly estimates semantic classifications and physical properties by combining visual and tactile data into a single semantic mapping framework. 
    RGB-D images are used to build a metric-semantic map that iteratively estimates semantic labels. 
    A property measurement is taken which in turn updates both the semantic class predictions and physical property estimates. 
    In the depicted example, the robot is unsure if the terrain in front of it is snow or ice from vision measurements alone (prior estimates) which dramatically affects the coefficient of friction and the associated gait that can be applied to safely traverse the terrain.
    The robot uses a tactile sensor attached to its manipulator to update its coefficient of friction estimation (posterior estimates), which then enables it to change gaits to cross the ice safely.}
    \label{fig:main}
\end{figure}

This paper introduces a multi-modal semantic prediction and property estimation framework to jointly estimate semantic labels alongside physical properties as depicted in Figure \ref{fig:main}.
The key insight is that a semantic class may be conditioned on its physical properties and physical properties may likewise be conditioned on the semantic class.
In this way, vision-based semantic classifications can be used to build a measurement model for physical properties, and tactile-based sensing can be used to update the semantic predictions via property measurements.
To apply this insight, this paper demonstrates how to recursively estimate physical properties from semantic class predictions using only visual information.
Subsequently, this paper describes how to exploit a tactile sensing modality to update the semantic classification and property estimates probabilistically and in closed-form using Bayesian inference.

The efficacy of the proposed approach is demonstrated on several hardware platforms including a robotic manipulator and a legged robot.
First, we show that by leveraging friction measurements obtained through tactile sensing we can improve semantic classification performance when compared to state-of-the-art semantic mapping methods that rely solely on visual information.
Second, we demonstrate that our approach estimates physical properties more accurately than a method using vision alone.
Multiple sensing modalities are used to estimate the coefficient of friction of the terrain surrounding a robot, which enables the use of an online risk-aware planner that switches the legged robot from a dynamic gait to a static, stable gait if the expected value of the coefficient of friction of the terrain falls below a given threshold.
Importantly, the proposed multi-modal framework allows the legged robotic system to complete a challenging terrain traversal task whereas state-of-the-art vision-based traversability estimation methods fail to accurately predict the traversability of the terrain resulting in the robot slipping and falling.
Finally, we show the broad applicability of the proposed framework for affordance-based property prediction by demonstrating door opening with unknown push-pull interactions.

The contributions of this work are two-fold: first, a technique to apply conjugate prior theory to enable the joint filtering of data collected from visual and tactile sensing modalities for semantic classification and property estimation; and second, an approximate conjugate prior for the Gaussian Mixture distribution that enables computationally tractable filtering.
This enables closed-form Bayesian updates given visual semantic classifications and tactile property measurements.
We demonstrate that by leveraging this probabilistic approach, we are able to improve semantic prediction performance across multiple metrics in simulation using only one property measurement.
These results are validated in hardware demonstrations.
Two case studies are subsequently presented that illustrate the utility of performing this type of multi-modal property estimation.
An open-source C++ and ROS package for the robot-agnostic implementation of the proposed approach as well as videos of the case studies are provided on the project page website\footnote{\href{https://roahmlab.github.io/multimodal_mapping/}{https://roahmlab.github.io/multimodal\_mapping/}}.
To the best of our knowledge, this is the first application of conjugate priors for fusing visual and tactile sensing modalities for semantic classification and property estimation.

The remainder of this paper is organized as follows: Section \ref{sec:rel_works} summarizes the semantic prediction and property estimation literature and Section \ref{sec:prelim} reviews preliminary material used throughout the paper.
Section \ref{sec:algorithm} provides an overview of our method.
Section \ref{sec:vision} introduces the vision-based semantic prediction pipeline and Section \ref{sec:tactile} the tactile-based semantic prediction and property estimation pipeline.
Section \ref{sec:implementation} describes the implementation.
Sections \ref{sec:simulation} and \ref{sec:experiments} describe the evaluation of our algorithm in simulation and using hardware demonstrations, respectively.
Section \ref{sec:conclusion} discusses future work and provides concluding remarks.

\section{Related Works} \label{sec:rel_works}
This section describes related work in semantic prediction and property estimation as well as the gap in the literature this work addresses.

\subsection{Semantic Prediction} \label{subsec:sem_pred_lit}

State-of-the-art methods for semantic classification almost exclusively rely on neural networks to produce pixel-wise semantic predictions \cite{guo2018review, hao2020brief, strudel2021segmenter}.
Often these approaches use task-dependent object designations as semantic labels \cite{deng2009imagenet, lin2014microsoft} such as abstract topological information \cite{kuipers1991robot} or material classifications \cite{upchurch2022materials}.
Unfortunately, semantic prediction methods often have no temporal consistency, meaning two subsequent images in a video stream may have vastly different semantic predictions.

To overcome this challenge and relate semantic predictions spatially and temporally, early work in semantic mapping projected semantic predictions onto a geometric representation forming a metric-semantic map \cite{kostavelis2015semantic}.
In this way, image-based semantic predictions have a shared frame of reference within this map.
These early methods employed voting-like schemes, regularization, and other algorithmic methods to fuse semantic predictions \cite{mccormac2017semanticfusion, sunderhauf2017meaningful}. 

Recent work in semantic mapping treats the semantic fusion process probabilistically to recursively estimate semantic labels \cite{gan2022multitask, wilson2023convbki}, building off of prior probabilistic approaches to binary occupancy mapping \cite{hornung13auro}.
These methods treat pixel-wise semantic predictions as measurements and probabilistically fuse these measurements over the map using Bayesian inference.
By approaching semantic prediction in this way, these approaches are robust to noisy pixel-wise semantic classifications and have been shown to be more accurate than methods that rely on image-based classifications alone \cite{ewen2022these}.

Multi-modal semantic prediction, and the corresponding semantic mapping extension, have been around since the rise of modern computer vision \cite{couprie2013indoor}.
Much of this work comes from the vision community and, as such, many of these methods rely on different vision-based modalities such as depth \cite{wang2015towards}, thermal imaging \cite{sun2020fuseseg}, and scale-invariant feature transform (SIFT) factors \cite{silberman2012indoor}.
Many modern multi-modal semantic classification methods are learning-based, where the multi-modal visual features are combined either at the input of or downstream within neural networks to output semantic classifications \cite{kim2018season, valada2020self}.

\subsection{Property Estimation} \label{subsec:prop_est_lit}

While semantic classification methods have seen widespread use in the robotics community, there are many robotic tasks (e.g., footstep planning, grasp planning, manipulation) where the underlying physical properties of the semantic classes are critical. 
Additionally, if semantic classifications are incorrect, the resultant property estimates may also be incorrect, leading to task failure.

Previous work in property estimation has focused on estimating the expected value for the coefficient of friction from vision alone \cite{brandao2016friction, noh2021surface, wang2019road}.
These methods have been extended in a recent paper that estimates probability distributions for the coefficient of friction given visual semantic class estimates by conditioning properties on semantic classes \cite{ewen2022these}.

Other methods for property estimation focus specifically on estimating traversability \cite{gan2022multitask, papadakis2013terrain, shneier2006performance}.
This is done to bypass the need to estimate terrain properties by instead estimating the regions in the environment a robot can traverse using visual features \cite{frey2022locomotion}.
This is often an ill-posed problem as traversability is not only terrain-dependent, but also dependent on the motion of the robot (e.g., walking vs. running) \cite{kim2006traversability}.
For example, it is possible for small insects to walk across a large body of water, but if a large robot applied the same gait it would most likely sink.

Notably, there has been a recent shift towards multi-modal property estimation in the literature.
One such approach proposes an online, self-supervised learning framework to estimate traversability by back-projecting previously traversed regions into the camera frame and training a network in a self-supervised manner to correctly label these previously seen regions as traversable \cite{frey2023fast}.
This method is considered multi-modal because it incorporates an implicit goal-success signal, in the form of the robot having not fallen over, alongside visual information.
Other work aims to encode property features in a latent space using tactile and visual information to train reinforcement learning agents \cite{sferrazza2023power}.
Recently, \cite{margolis2023learning} proposed a method for learning motion policies by training a network to relate image-based features and physical properties such as friction in simulation.
It was shown that the trained network accurately predicted properties in hardware demonstrations, however these predictions only output a single scalar value for properties without considering uncertainty in the prediction.
A similar framework for multi-modal probabilistic property estimation within a semantic mapping framework was also demonstrated for the purposes of traversability estimation \cite{cai2023evora}.
In this case, the terrain classes are modelled probabilistically using Dirichlet distributions (analogous to \cite{ewen2022these, gan2022multitask}) and trained networks are then used to predict the likelihood of slipping using data collected offline.

\subsection{Gap in the Literature}

The methods discussed in the prior two subsections have two notable shortcomings. 
The first is that, while existing property estimation algorithms condition physical properties on semantic classifications, this conditioning of properties is immutable, meaning the property values of semantic classes cannot be updated during operation \cite{cai2023evora, ewen2022these, gan2022multitask}.
This is undesirable especially when such physical properties have intra-class variation or there are errors in these initial estimates.
For example, the coefficient of friction of ice depends on the conditions of ice formation, outdoor temperature, and the presence of snow on the surface of the ice.
Assuming that the coefficient of all ice is identical may result in unsafe or overly conservative behavior.

The second short-coming of these methods is the one-way conditioning of physical properties on semantic classes.
This can be unduly limiting because visual information alone may not be sufficiently informative. 
For example, in Fig. \ref{fig:main} it is challenging for the semantic segmentation network to differentiate between snow and ice. 
We note that in this work we make the assumption that tactile information may be used to disambiguate visual uncertainty.
Using a tactile sensor in this instance to estimate the friction coefficient can inform and improve the accuracy of the semantic classification. 
Using physical properties to inform semantic classification can enable the application of multiple sensing modalities while providing a means of verification for semantic predictions output by trained networks.
This improves the accuracy of semantic predictions without requiring new training data and network retraining.

To address these short-comings, this paper proposes a method to \textit{jointly} estimate semantic classifications and physical properties online using Bayesian inference.
Semantic classifications from vision enable the prediction of physical properties, and tactile measurements enable the online update of physical property estimates. 
This in turn enables the semantic classification likelihoods to be updated.

\section{Preliminaries} \label{sec:prelim}
This section introduces notation, moments, and conjugate pairs.
Vectors, written as columns, are typeset in bold and lowercase, while sets and matrices are typeset in uppercase. 
The element $i$ of a vector $\boldsymbol{x}$ is denoted as $x_i$.
An n-dimensional closed interval is denoted by $[a, b]^n$.
Let $p$ denote a normalized probability mass or density function. 
Following the convention of \cite{thrun2002probabilistic}, we distinguish between different probability mass or density functions by the variable used as an input argument into each $p$.
To avoid confusion, the term \textit{multi-modal} refers to multiple sensing modalities, while \textit{mixture} refers to a distribution with more than one mode.
Likewise, we use \textit{vision} to refer to RGB-D data.

\subsection{Conjugate Pairs} \label{subsec:conjugate_pairs}
In probability, conjugate distributions are a pair of distributions such that, given a prior and a likelihood, the posterior computed using Bayes' theorem belongs to the same family of distributions as the prior \cite{diaconis1979conjugate}.
This enables tractable computation of closed-form solutions for the posterior \cite{anderson2012optimal}.
We introduce two such conjugate pairs here: the Dirichlet and Categorical distributions, and the Dirichlet Normal-Inverse-Gamma product and Gaussian mixture distributions.

\subsubsection{Dirichlet Conjugate Prior} \label{subsubsec:first_conj_pair}
Let $z \in  \{1,\ldots,k\}$ denote a discrete random variable.
The probability mass function of the Categorical distribution represents the probability that a sample $z$ belongs to class $i$:
\begin{equation} \label{eq:categorical}
    p(z = i | \boldsymbol{\theta}) = Cat(z = i | \boldsymbol{\theta}) = \theta_i,
\end{equation}
where $i \in \{1, 2, \dots, k\}$, $\boldsymbol{\theta} \in [0, 1]^k$, and $\sum_i \theta_i = 1$.

The Dirichlet distribution is a continuous k-variate probability distribution that is parameterized by a vector $\boldsymbol{\alpha} \in \mathbb{R}^k_{\geq 0}$ with probability density: 
\begin{align} \label{eq:dirichlet}
    p(\boldsymbol{\theta} | \boldsymbol{\alpha}) &= Dir(\boldsymbol{\theta} | \boldsymbol{\alpha}) \\
    &= \frac{\Gamma(\sum_{j=1}^k \alpha_j)}{\sum_{j=1}^k \Gamma(\alpha_j)} \prod_{j=1}^k \theta_j^{\alpha_j-1},
\end{align}
where 
\begin{equation}
    \Gamma(\alpha_j) = \int_0^\infty x^{\alpha_j-1} \exp(-x) dx.
\end{equation}
The Dirichlet distribution is a conjugate prior to the Categorical distribution as is formalized in the following theorem  whose proof is provided in Appendix \ref{app:A}:

\begin{thm} \label{thm:dir_cat_conj}
Let $\mathcal{Z} = \{z_1, \dots, z_n\}$ be a set of measurements drawn from a Categorical distribution, $p(z_j = i | \boldsymbol{\theta})$, and let the prior for $\boldsymbol{\theta}$ be a Dirichlet distribution, $p(\boldsymbol{\theta}|\boldsymbol{\alpha})$.
The posterior computed using Bayes' theorem, $p(\boldsymbol{\theta}| \mathcal{Z}, \boldsymbol{\alpha})$, is also a Dirichlet distribution such that:
\begin{equation} \label{eq:dir_posterior}
    p(\boldsymbol{\theta}| \mathcal{Z}, \boldsymbol{\alpha}) = Dir(\boldsymbol{\theta}| \Tilde{\boldsymbol{\alpha}}),
\end{equation}
where
\begin{align} \label{eq:alpha_update}
    \tilde{\alpha}_j &= \alpha_j + \sum_{z_i \in \mathcal{Z}} 1\{z_i = j\},
\end{align}
and $1\{z_i = j\}$ is equal to $1$ when the expected class of measurement $z_i$ is class $j$ and is zero otherwise.
\end{thm}

Theorem \ref{thm:dir_cat_conj} enables the computation of the predictive posterior.
In particular, one of the goals of this paper is to predict the probability that a new measurement belongs to class $i$ given prior measurements $\mathcal{Z}$:
\begin{align} \label{eq:bayes_integral}
    p(z = i| \mathcal{Z},\boldsymbol{\alpha}) &= p(z = i| \Tilde{\boldsymbol{\alpha}}) \\
    &= \int_{\boldsymbol{\theta}} p(z = i|\boldsymbol{\theta}) p(\boldsymbol{\theta}|\Tilde{\boldsymbol{\alpha}}) \text{d}\theta,
\end{align}
where $ p(z = i|\boldsymbol{\theta})$ represents the Categorical likelihood and $p(\boldsymbol{\theta}|\Tilde{\boldsymbol{\alpha}})$ the Dirichlet posterior probability. 
Computing this integral exactly is challenging.
Fortunately, the theory of conjugate priors enables a closed-form solution  {\cite[(3)]{tu2014}}:
\begin{equation} \label{eq:dir_to_cat}
    p(z = i| \tilde{\boldsymbol{\alpha}}) = \frac{\tilde{\alpha}_i}{\sum_{j=1}^k \tilde{\alpha}_j}.
\end{equation}

\subsubsection{Dirichlet Normal-Inverse-Gamma Conjugate Prior} \label{subsubsec:second_conj_pair}
Let $\psi \in \R$ denote a continuous random variable.
A Gaussian mixture is a continuous probability density function
\begin{equation} \label{eq:gauss_mix_model}
    p(\psi | \Theta) = \sum_{i=1}^k w_i \N (\psi | \mu_i, \sigma_i^2),
\end{equation}
where $\Theta = \{w_i, \mu_i, \sigma_i^2\}_{i=1}^k$ and $\sum_{i=1}^k w_i = 1$.
Next, we present a candidate for the conjugate prior of the Gaussian mixture model by conditioning the parameters, $\Theta$, on a set of hyperparameters, $\Psi$.
Previously, the Dirichlet distribution was used to parameterize these variables to form a conjugate pair.
We perform a similar trick here using the Dirichlet Normal-Inverse-Gamma product distribution as the prior on $\Theta$.

The Dirichlet Normal-Inverse-Gamma product distribution is defined as 
\begin{equation} \label{eq:dir_normal_gamma}
    p(\Theta | \Psi) = Dir(\boldsymbol{w} | \boldsymbol{a}) \prod_{i=1}^k \NIG(\mu_i, \sigma^2_i | \tau_i, \kappa_i, \beta_i, \gamma_i),
\end{equation}
where $\Psi = \{a_i, \tau_i, \kappa_i, \beta_i, \gamma_i\}_{i=1}^k$, $\kappa_i, \beta_i, \gamma_i > 0$, and the Normal-Inverse-Gamma distribution is defined as
\begin{align}
    \NIG(\mu_i, \sigma_i^2 | \tau_i, \kappa_i, \beta_i, \gamma_i) &= \frac{\sqrt{\kappa_i}}{\sqrt{2 \pi \sigma_i^2}} \frac{\gamma_i^{\beta_i}}{\Gamma(\beta_i)} \frac{1}{\sigma_i^2}^{\beta_i+1} \cdot \\
    &\cdot \exp(-\frac{2\gamma_i + \kappa_i(x-\tau_i)^2}{2 \sigma_i^2}). \nonumber
\end{align}
When a measurement $\psi$ is taken, we assume it is drawn from the Gaussian mixture measurement likelihood.
The posterior of $\Theta$ is then computed using the following theorem from {\cite[(6)]{jaini2016online}}:

\begin{thm} \label{thm:true_posterior}
    Let $\psi$ be a measurement drawn from a Gaussian mixture $p(\psi | \Theta)$.
    Let the prior for $\Theta$ be a Dirichlet Normal-Inverse-Gamma product distribution.
    Then the posterior computed using Bayes' theorem, $p(\boldsymbol{\Theta}| \psi, \Psi)$, is:
    \begin{align} \label{eq:conj_post}
        p(\Theta | \psi, \Psi) = \frac{1}{M}\sum_{j=1}^k & c_j Dir(\boldsymbol{w} | \tilde{\boldsymbol{a}}_j) \cdot \nonumber\\
        &\cdot \NIG(\mu_j, \sigma^2_j | \Tilde{\tau}_j, \Tilde{\kappa}_j, \Tilde{\beta}_j, \Tilde{\gamma}_j) \cdot \nonumber\\ 
        &\cdot \prod_{i \neq j}^k \NIG(\mu_i, \sigma^2_i | \tau_i, \kappa_i, \beta_i, \gamma_i),
    \end{align}
    where
    \begin{align} \label{eq:true_posterior}
        \tilde{a_j} &= a_j +1, \\
        \tilde{\tau_j} &= \frac{\kappa_j \tau_j + \psi}{\kappa_j +1}, \\
        \tilde{\kappa_j} &= \kappa_j +1, \\
        \tilde{\beta_j} &= \beta_j +\frac{1}{2}, \\
        \tilde{\gamma_j} &= \gamma_j + \kappa_j \frac{(\psi - \tau_j)^2}{2(1+\kappa_j)}, \\
        c_j  &= \sqrt{\frac{\kappa_j}{\tilde{\kappa_j}}} \frac{\Gamma(\tilde{\beta}_j)}{\Gamma(\beta_j)}\frac{{\gamma_j}^{(\beta_j)}}{{\tilde{\gamma}_j}^{(\tilde{\beta}_j)}},\label{eq:true_posterior_end}
    \end{align}
    $\tilde{\boldsymbol{a}}_j = \{a_1, \ldots a_{j-1}, \Tilde{a}_j, a_{j+1}, \ldots, a_k \}$  and $M$ is the normalizing factor.
\end{thm}
Using Theorem \ref{thm:true_posterior}, note the posterior, \eqref{eq:conj_post}, is not in the same family of distributions as the prior, \eqref{eq:dir_normal_gamma}.
More troublingly, the number of terms in \eqref{eq:conj_post} grows exponentially with the number of measurements.
Rather than apply Theorem \ref{thm:true_posterior}, we implement the method of moments described in the next section to approximate \eqref{eq:conj_post} as a Dirichlet Normal-Inverse-Gamma product distribution.

\subsection{Method of Moments} \label{subsec:moment_matching}

Let $\boldsymbol{\lambda} = \{\lambda_1, \ldots, \lambda_n\}$ be an n-dimensional continuous random variable with probability density function $p(\boldsymbol{\lambda} | \xi)$ conditioned on parameters $\xi$.
The $i$-th order moment is defined as
\begin{equation} \label{eq:moments}
    M_{g_i(\boldsymbol{\lambda})}(p) = \E [g_i(\boldsymbol{\lambda})]
\end{equation}
where $g_i$ is a monomial of $\boldsymbol{\lambda}$ of degree $i$.
For some distributions, there exists a finite set of such moments, termed the \textit{sufficient moments} \cite{jaini2016online} that fully define the set of parameters $\xi$.
This means that one can construct the probability density function for this distribution by just using the sufficient moments.
We denote the set of sufficient moments of a distribution $p$ as $\mathbb{S}_p$.

The method of moments is a technique that fits a distribution to a set of sampled data by matching the sufficient moments of the distribution with the empirical moments computed from data.
This approach may also be used to approximate a distribution by another distribution in a different family of distributions by matching sufficient moments of the first distribution with those of the second distribution.
For our purposes, we use the method of moments to approximate the posterior computed using Theorem \ref{thm:true_posterior} with a probability density from the family of Dirichlet Normal-Inverse-Gamma product distributions.
This may be thought of as projecting the first distribution onto the family of the second distribution \cite{jaini2016online}.
This process is summarized in Algorithm \ref{alg:moments} for projecting \eqref{eq:conj_post} onto the family of distributions of \eqref{eq:dir_normal_gamma}. 

\begin{algorithm}[t]
\DontPrintSemicolon

\BlankLine

\Requires{Dataset $\mathcal{D}=\{\psi_l\}_{l=1}^N$, Prior $p(\Theta | \Psi)$}
\For{$\psi \in \mathcal{D}$}{
    Compute $p(\Theta | \psi, \Psi)$\tcp*{\eqref{eq:conj_post}}
    $\E[g_j(\boldsymbol{\theta})] \leftarrow$ moments($p(\Theta | \psi, \Psi)$) \tcp*{\eqref{eq:moments}}
    $\hat{\Psi} \leftarrow$ matchMoments() \tcp*{\eqref{eq:first_hat}-\eqref{eq:last_hat}}
    Compute $p(\Theta | \hat{\Psi})$
}
\caption{Method of Moments} \label{alg:moments}
\end{algorithm}

\subsection{Revisiting the Dirichlet Normal-Inverse-Gamma Product}

By implementing Algorithm \ref{alg:moments}, we project \eqref{eq:conj_post} onto the family of Dirichlet Normal-Inverse-Gamma product distributions resulting in an approximate posterior.
To compute this projection using the method of moments, we first need the sufficient moments of the Dirichlet Normal-Inverse-Gamma product distribution.
From \cite[\S 4.1]{jaini2016online}, these sufficient moments are $\mathbb{S}_p = \{\mu_i, \sigma^2_i, \sigma_i^4, \mu_i^2 \sigma^2_i, w_i, w_i^2\}_{i=1}^k$.
This process is summarized in the following theorem whose proof is given in Appendix \ref{app:B}:

\begin{thm} \label{thm:bmm_conj}
Let $\psi$ be a measurement drawn from a Gaussian mixture $p(\psi | \Theta)$.
Let the prior for $\Theta$ be a Dirichlet Normal-Inverse-Gamma product distribution.
Then the posterior computed using Bayes' theorem, $p(\boldsymbol{\theta}| \psi, \Psi)$, and projected onto the family of Dirichlet Normal-Inverse-Gamma product distributions via Algorithm \ref{alg:moments} yields:
\begin{equation} \label{eq:bmm}
p(\Theta |\hat{\Psi}) = Dir(\boldsymbol{w} | \hat{\boldsymbol{a}}) \prod_{i=1}^k \NIG(\mu_i, \sigma^2_i | \hat{\tau}_i, \hat{\kappa}_i, \hat{\beta}_i, \hat{\gamma}_i).
\end{equation}
The parameters $\hat{\Psi}$ are computed using the sufficient moments of \eqref{eq:conj_post}:
\begin{align}
    \hat{\tau_i} & = \mathbb{E}[\mu_i], \label{eq:first_hat} \\
    \hat{\kappa_i} &= \frac{1}{\mathbb{E}[\mu_i^2 \sigma^2_i] - \mathbb{E}[\mu_i]^2\mathbb{E}[\sigma^2_i] }, \\
    \hat{\beta_i} & = \frac{\mathbb{E}[\sigma^2_i]^2}{\mathbb{E}[\sigma_i^4] - \mathbb{E}[\sigma^2_i]^2}, \\
    \hat{\gamma_i} & = \frac{\mathbb{E}[\sigma^2_i]}{\mathbb{E}[\sigma_i^4] - \mathbb{E}[\sigma^2_i]^2}, \\
    \hat{a_i} & = \mathbb{E}[w_i] \frac{\mathbb{E}[w_i] -\mathbb{E}[w_i^2]}{\mathbb{E}[w_i^2] -\mathbb{E}[w_i]^2}. \label{eq:last_hat}
\end{align}
\end{thm}
Theorem \ref{thm:bmm_conj} is applied sequentially when multiple measurements are given.
We summarize this procedure in Algorithm \ref{alg:moments}.
The posterior predictive distribution is then given as $p(\psi | \hat{\Theta})$, which is a Gaussian mixture, where 

\begin{equation}
    \hat{\Theta} = \E[p(\Theta | \hat{\Psi})].
\end{equation}

\begin{figure*}
    \centering
    \includegraphics[width=2\columnwidth]{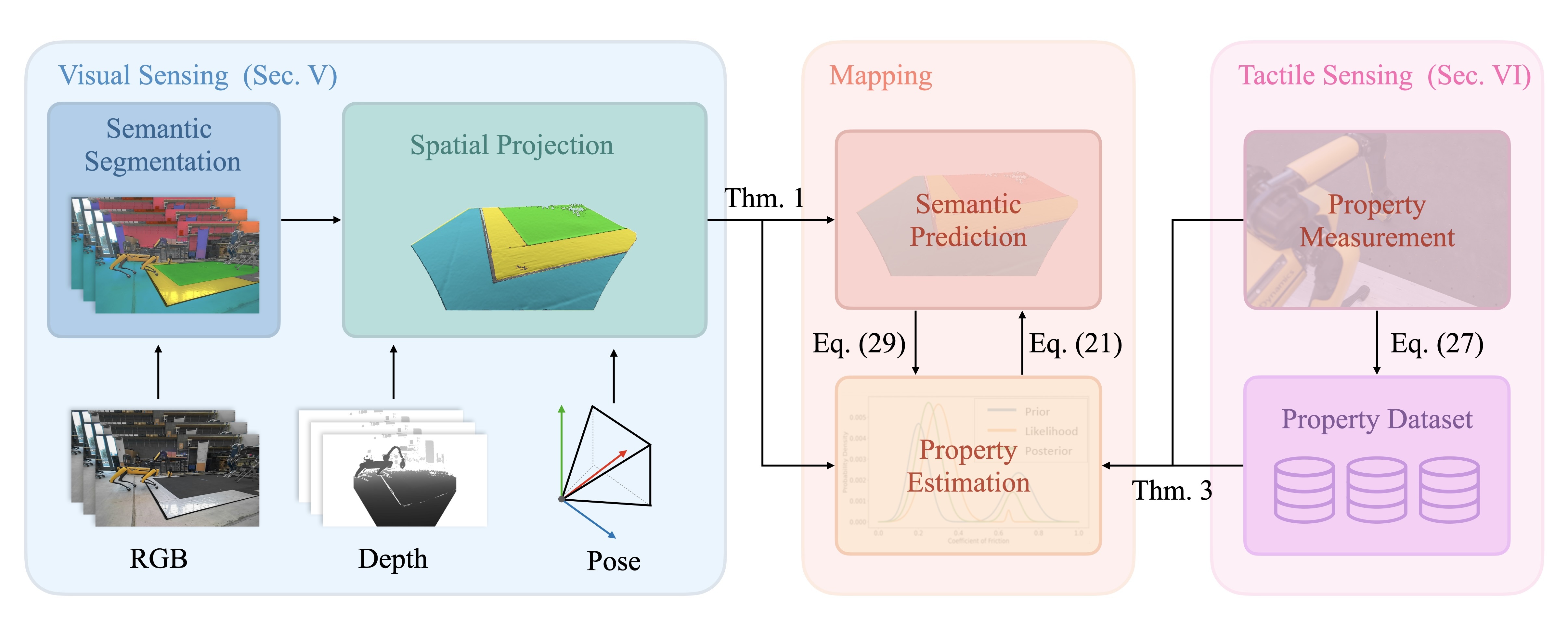}
     \caption{A flow diagram illustrating Algorithm \ref{alg:pred}.
     A semantic classification algorithm predicts pixel-wise classes from RGB images that are then projected into a common mapping frame using the aligned depth image, camera intrinsics, and estimated camera pose. 
     This semantic point cloud is used to build a metric-semantic map. 
     When a property measurement is taken, Algorithm \ref{alg:moments} is used to update the semantic and property estimates.}
     \label{fig:algorithm_overview}
\end{figure*}

\section{Algorithm Overview} \label{sec:algorithm}
\begin{algorithm}[t]
\DontPrintSemicolon
$\mathcal{M} \leftarrow$ initMap() \label{alg:line:initMap} \;
$\Theta \leftarrow$ initUncertainty() \label{alg:line:initUncertainty} \;
 \While{robot is running}
 {
 $s \leftarrow$ getSemanticPointCloud()  \label{alg:line:pointcloud} \;
 $\mathcal{M} \leftarrow$ updateMap($s$) \label{alg:line:updateMap} \tcp*{Sec.\ref{sec:vision}}
 $\Theta \leftarrow$ updateSemanticUncertainty($s$) \label{alg:line:likelihood} \tcp*{Sec.\ref{sec:vision}}
 \If{ user requests property measurement taken}
 {
 $\psi \leftarrow$ getMeasurement() \label{alg:line:measurement} \;
 $\mathcal{V} \leftarrow$ getMeasurementLocation() \label{alg:line:location} \tcp*{Sec.\ref{sec:experiments}}
 $\Theta \leftarrow$ measurementUpdate($\psi, \mathcal{V}$) \label{alg:line:update} \tcp*{Alg.\ref{alg:moments}}
 }
}
\caption{Semantic and Property Prediction} \label{alg:pred}
\end{algorithm}

The proposed algorithm takes as input a stream of RGB-D images and physical property measurements from a tactile sensor and outputs semantic classifications and physical property estimates which are modelled probabilistically.
We provide an overview of the proposed method presented in Algorithm \ref{alg:pred}.
We focus on tactile and visual modalities, however our method may be adapted to different sensing modalities in a straight-forward manner.

A metric-semantic map and Dirichlet parameters are first initialized, where $\boldsymbol{\alpha} = \boldsymbol{1}_{k \times 1}$ (Lines \ref{alg:line:initMap} - \ref{alg:line:initUncertainty}). 
We assume that, while the robot is running, it is collecting exteroceptive data in the form of RGB-D images or lidar data.
Using an off-the-self semantic segmentation network of choice, this data is transformed into a semantic point cloud \cite{nixon2012} (Line \ref{alg:line:pointcloud}).
The geometric information of this point cloud is used to update the metric map (Line \ref{alg:line:updateMap}) while the semantic classifications are used to compute the semantic class posterior (Line \ref{alg:line:likelihood}), as described in Section \ref{sec:vision}.

Locations for property measurements are specified by the user.
When a property measurement is taken using a tactile sensor, we first compute the region over which the measurement was taken within the map and use moment matching (Algorithm \ref{alg:moments}) to update the semantic classifications and property estimates within this region (Lines \ref{alg:line:measurement}-\ref{alg:line:update}).
We describe this step in more detail in Section \ref{sec:tactile}.
A flow diagram illustrating this procedure is given in Figure \ref{fig:algorithm_overview}.

\section{Vision-Based Estimation} \label{sec:vision}
This section describes how to implement Bayesian inference to recursively estimate semantic classifications using visual data.
This corresponds to the \textit{Visual Sensing} block in Figure \ref{fig:algorithm_overview}.
First, we discuss how to obtain a semantic segmentation point cloud and then discuss how to probabilistically fuse this data within a geometric representation.

Semantic segmentation assigns  task-dependent class probability scores to each pixel in an image.
It is commonplace to use neural networks to estimate pixel-wise class probability scores \cite{garcia2018}.
Let $I \in \mathbb{R}^{w \times h \times 3}$ denote an RGB image, where $h,w$ are the height and width of the image in pixels.
A trained semantic segmentation network takes an input image $I$ and outputs the pixel-wise class probability scores $T \in \mathbb{R}^{w \times h \times k}$ for $k$ semantic classes in the form of a k-dimensional Categorical distribution.
By employing a one-hot encoding over these pixel-wise scores, $H = 1\{T\}$ such that $H \in \R^{w \times h}$, we may interpret the network as drawing pixel-wise class samples from pixel-wise Categorical distributions.
We denote these as the \textit{semantic class measurements}, the accuracy of which depends on the network used.

We use an aligned depth image, camera intrinsics, and camera pose estimate to project the semantic class measurements $H$ into a fixed global coordinate frame \cite[(10.38)]{nixon2012} to obtain a point cloud representation of the semantically segmented image.
Akin to \cite{xiang2017rnn} we implement a modified version of Kinect Fusion \cite{izadi2011kinectfusion}, a truncated signed distance field (TSDF), as the geometric representation onto which semantic class measurements are projected.
Within each voxel we store unique Dirichlet parameters $\boldsymbol{\alpha}$.
Given new semantic class measurements, we update these parameters in each voxel using \eqref{eq:alpha_update}.
Thus, semantic classification estimates may be recursively obtained with uncertainty estimates via \eqref{eq:dir_posterior}.

\section{Tactile-Based Estimation} \label{sec:tactile}
This section describes how semantic class estimates may be used to predict physical properties and how we use measurements of these properties to update semantic class likelihoods.
This corresponds to the \textit{Tactile Sensing} block in Figure \ref{fig:algorithm_overview}.
Our goal is to compute a measurement likelihood for physical properties using semantic class likelihoods introduced in Section \ref{sec:vision}.

Motivated by \cite{ewen2022these, nguyen2021}, we construct a conditional probability distribution $ p(\boldsymbol{\psi} \mid \mathcal{Z})$ for a property $\psi$ with semantic data $\mathcal{Z}$.
By applying the law of total probability and including the $\boldsymbol{\alpha}$ parameters introduced in Section \ref{subsubsec:first_conj_pair}, we derive the following expression:
\begin{equation} \label{eq:contitionalindep_infer}
        p(\boldsymbol{\psi} \mid \mathcal{Z}, \boldsymbol{\alpha}) = \sum_{i = 1}^k p(\boldsymbol{\psi} \mid z = i) p(z = i \mid \mathcal{Z}, \boldsymbol{\alpha}).
\end{equation}
There are two components to this equation.
The first, $p(z = i \mid \mathcal{Z}, \boldsymbol{\alpha})$, is the posterior predictive likelihood of a region of the map belonging to semantic class $i$ given prior semantic class measurements within that region.
This posterior predictive likelihood was derived in Section \ref{subsec:conjugate_pairs} and is given by \eqref{eq:dir_to_cat}.

Physical properties are not constant across a semantic class and thus should be estimated via a distribution of possible values, not by a single value.
The second component, $p(\boldsymbol{\psi} \mid z = i)$, denotes this likelihood distribution and is approximated as a Gaussian distribution \cite{ewen2022these}.
Substituting \eqref{eq:dir_to_cat} and the formula for the class-wise Gaussian model into \eqref{eq:contitionalindep_infer} gives a closed-form estimate for the physical properties given semantic class measurements:
\begin{equation} \label{eq:property_comp}
    p(\boldsymbol{\psi} \mid \mathcal{Z}, \boldsymbol{\alpha}) = \sum_{i=1}^k \frac{\alpha_i}{\sum_{j=1}^k \alpha_j} \mathcal{N}(\mu_i, \sigma_i^2).
\end{equation}
Note that this model is equivalent to the Gaussian mixture model presented in \eqref{eq:gauss_mix_model} where the weights are computed using the predictive posterior of the semantic classifications.
We use this as the measurement likelihood for the properties of interest.

\begin{table}[!t]
\centering
\begin{tabular}{rcc}
\toprule
\multirow{2}{*}{\textbf{Semantic Class}} & \multicolumn{2}{c}{Gaussian Parameters} \\ & \multicolumn{1}{c}{\; \; \; $\mu$ \; \; \;} & $\sigma$ \\
\midrule 
Concrete & \multicolumn{1}{c}{0.543} & 0.065 \\
Grass & \multicolumn{1}{c}{0.577} & 0.077 \\
Rock & \multicolumn{1}{c}{0.478} & 0.133 \\
Wood & \multicolumn{1}{c}{0.372} & 0.055 \\
Rubber & \multicolumn{1}{c}{0.616} & 0.048 \\
Plastic & \multicolumn{1}{c}{0.311} & 0.045 \\
Snow & \multicolumn{1}{c}{0.390} & 0.071 \\
Ice & \multicolumn{1}{c}{0.192} & 0.046 \\
\bottomrule
\end{tabular}
\caption{Gaussian parameters for the coefficient of friction of semantic classes. These values are computed using the material friction dataset provided by \cite{ewen2022these} and assume rubber as the contact material.}
\label{table:friction}
\end{table}

Given that the Gaussian mixture model is used as the measurement likelihood, we use the approximate conjugate prior of this distribution, the Dirichlet Normal-Inverse-Gamma product, to represent uncertainty in the semantic classifications, $w_i$, and class-wise property parameters, $\mu_i$ and $\sigma^2_i$.
Note that the Gaussian mixture weights, $w_i$, are computed using the Dirichlet parameters $\boldsymbol{\alpha}$.
After a property measurement is taken, Algorithm \ref{alg:moments} is implemented to update the belief over these parameters via \eqref{eq:bmm}. This algorithm is applied sequentially if multiple measurements are taken.

Using Algorithm \ref{alg:moments}, the semantic class likelihood, $w_i$, as well as the property parameters themselves, $\mu_i$ and $\sigma^2_i$, are updated.
Furthermore, the Dirichlet Normal-Inverse-Gamma product posterior provides a means of uncertainty quantification in the posterior estimates of the semantic class likelihoods and property values, unlike learning-based approaches.

\section{Implementation} \label{sec:implementation}
Algorithms \ref{alg:moments} and \ref{alg:pred} are implemented in C++ and include a ROS interface which will be released upon final submission of the paper.
Experiments are conducted on a laptop with a 5.4GHz i9-13900HX processor, 32GB of RAM, and an Nvidia RTX 4060 GPU. 

We use a custom implementation of the SegFormer network \cite{xie2021segformer} trained on the Dense Material Segmentation Dataset \cite{upchurch2022materials}.
The output of the network is then post-processed with a segment-based voting scheme using FastSAM \cite{frey2023fast} in a similar manner to \cite{chen2023semantic}.
When a property measurement is taken, Algorithm \ref{alg:moments} updates a region of the geometric representation segmented using FastSAM.
This incentivizes regions with spatial proximity and visual similarity to be updated using a single measurement rather than requiring one measurement for each voxel, making the algorithm more efficient. 

Table \ref{table:friction} is used to initialize the Gaussian parameters for the coefficient of friction for the relevant semantic classes.
The values in Table \ref{table:friction} were computed using the material friction dataset provided by \cite{ewen2022these}.
If a friction measurement is taken of a material class not present in Table \ref{table:friction}, Algorithm \ref{alg:moments} predicts the class from Table \ref{table:friction} with the most similar friction values.

\begin{figure*}
    \centering
    \includegraphics[width=2\columnwidth]{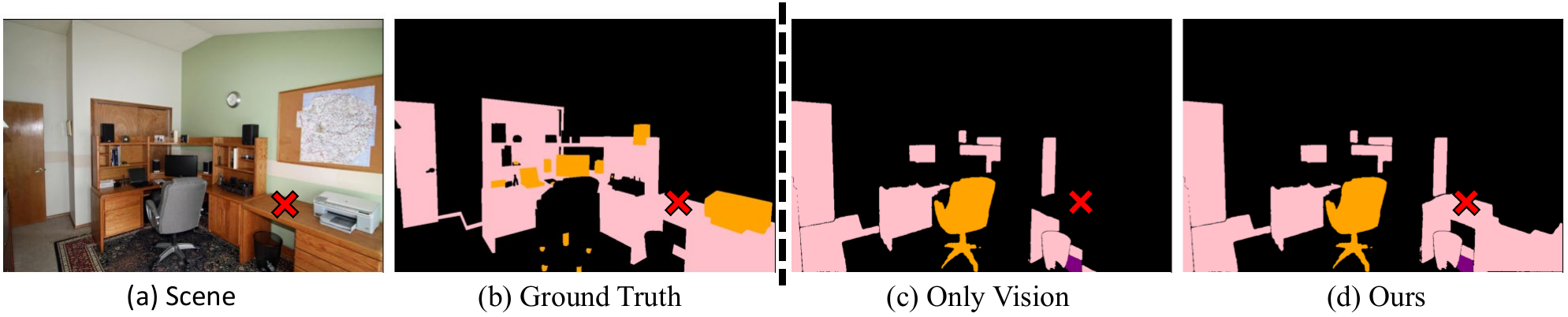}
     \caption{Results for a single simulated experiment. The image (a) and ground truth semantic labels (b) are from the Dense Material Segmentation Dataset \cite{upchurch2022materials}. The semantic segmentation predictions (c) do not classify parts of the desk as wood. A friction measurement is simulated using the ground-truth semantic label and Table \ref{table:friction} and is sampled from the pixel highlighted by the red cross in the RGB image. Algorithm \ref{alg:moments} then computes the correct posterior semantic label (d).}
     \label{fig:sim_demo}
\end{figure*}

For hardware demonstrations, static friction measurements are taken using a force-torque sensor which measures contact forces during the motion of an end-effector against a surface.
These forces are converted to friction measurements using the equation relating normal and tangential forces:
\begin{equation} \label{eq:friction}
    \psi = \frac{F_t}{F_n},
\end{equation}
where $F_t$ is the force tangential to the contact normal, $F_n$ is the force parallel to the contact normal, and $\psi$ is the coefficient of friction.
A low-pass filter is used to filter out measurement noise from the force-torque sensor.

\section{Simulation Results} \label{sec:simulation}
We validate our approach in simulation and demonstrate property measurements improve semantic predictions.
The simulation experiments also validate the approximation accuracy of the moment matching method in Algorithm \ref{alg:moments} in approximating the posterior via Theorem \ref{thm:bmm_conj}.

\subsection{Dense Material Segmentation Dataset Evaluation} \label{subsec:dmsd_eval}
We use 1800 images and ground-truth semantic labels taken from the testing set of the Dense Material Segmentation Dataset \cite{upchurch2022materials} and the pre-trained semantic segmentation neural network described in Section \ref{sec:implementation} is used to predict semantic labels.
Locations of incorrect semantic classification are determined by comparing to the ground-truth semantic classifications provided by the dataset.
Misclassified pixels in each image are then randomly selected and a friction measurement is drawn from a Gaussian distribution whose parameters are specified in Table \ref{table:friction} using the corresponding ground-truth semantic label.
The parameters and classes in Table \ref{table:friction} are identical to those in \cite{ewen2022these} which were validated through thorough real-world experimentation.
In particular, \cite{ewen2022these} found that a Gaussian distribution using the parameters in Table \ref{table:friction} modeled real-world measurements well.
Note that we use a similar prior model during our real-world experiments as are described in Section \ref{sec:experiments}.

The semantic prediction is used to initialize the Dirichlet parameters such that $\boldsymbol{\alpha} = 1_{k\times 1} + 1\{z = j\}$.
The Gaussian parameters are initialized via Table \ref{table:friction} and the remaining Dirichlet Normal-Inverse-Gamma parameters are initialized as $\boldsymbol{a} = \boldsymbol{\alpha}$, and $\tau_i = \mu_i$, $\kappa_i = 1$, $\beta_i = \sigma_i^2 / \tau_i$, $\gamma_i = \sqrt{\beta_i}/C$, and $C = 40$ for each semantic class.

\begin{figure*}
    \centering
    \includegraphics[width=2\columnwidth]{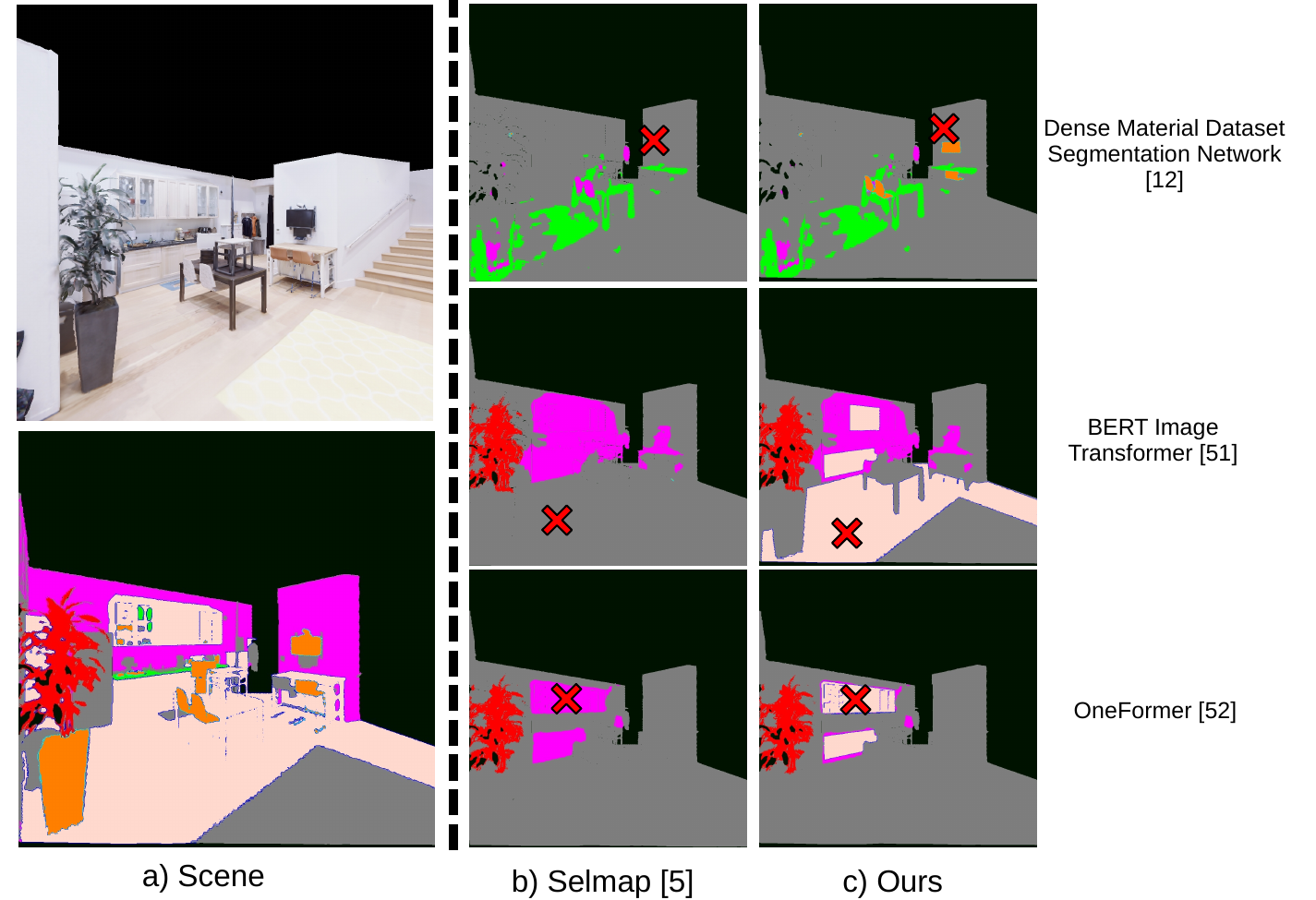}
     \caption{\reviewadd{Results for a single simulated experiment in the Habitat simulator \cite{savva2019habitat}. a) The input RGB image with ground truth semantic labels is shown. On the right-hand side are the outputs of the Selmap \cite{ewen2022these} baseline and our proposed approach using three different semantic segmentation networks. b) The incorrect semantic labelling is reflected in the Selmap implementation as the semantic segmentation networks all predict incorrect classifications in various regions in the scene. By exploiting tactile measurements taken at the locations denoted by the red 'X', c) our proposed approach is able to correct these erroneous semantic predictions.}}
     \label{fig:replica_demo}
\end{figure*}

After a simulated friction measurement is randomly sampled, Algorithm \ref{alg:moments} is invoked and the posterior predictive likelihood is computed.
The Segment Anything Model \cite{chen2023semantic} computes a mask over the input RGB image using the chosen pixel as the query point and the semantic predictions over this region are updated corresponding to the posterior predictive likelihood output by Algorithm \ref{alg:moments}.
An example of this computed posterior is demonstrated in Figure \ref{fig:sim_demo}.

\begin{table*}[!t]
\centering
\begin{tabular}{lcccccccccc}
\toprule
  & & Concrete & Grass & Rock & Wood & Rubber & Plastic & Snow & Ice & Total  \\
\midrule
\multirow{2}{*}{Accuracy ($\uparrow$)} & Vision Only & 0.70 & 0.74 & 0.70 & 0.76 & 0.81 & 0.82 & 0.92 & 0.82 & 0.78 \\
                          & Ours        & \textbf{0.83} & \textbf{0.77} & \textbf{0.79} & \textbf{0.82} & \textbf{0.83} & \textbf{0.85} & \textbf{0.99} & \textbf{0.84} & \textbf{0.83} \\
\bottomrule
\bottomrule\addlinespace[\belowrulesep]
\multirow{2}{*}{PSNR ($\uparrow$)} & Vision Only & 8.48 & 11.91 & 11.90 & 8.97 & 12.01 & 11.26 & 16.66 & 11.12 & 9.96 \\
                      & Ours        & \textbf{11.59} & \textbf{12.07} & \textbf{13.43} & \textbf{10.44} & \textbf{12.07} & \textbf{12.07} & \textbf{21.76} & \textbf{11.44} & \textbf{11.20} \\
\bottomrule
\bottomrule\addlinespace[\belowrulesep]
\multirow{2}{*}{SSIM ($\uparrow$)} & Vision Only & 0.70 & 0.73 & 0.69 & 0.74 & \textbf{0.79} & 0.80 & 0.92 & 0.81 & 0.76 \\
                      & Ours        & \textbf{0.80} & \textbf{0.75} & \textbf{0.77} & \textbf{0.79} & \textbf{0.79} & \textbf{0.82} & \textbf{0.98} & \textbf{0.82} & \textbf{0.80} \\
\bottomrule
\bottomrule\addlinespace[\belowrulesep]
\multirow{2}{*}{Binary Cross Entropy ($\downarrow$)} & Vision Only & 2.72 & 2.67 & 2.73 & 2.58 & 2.86 & 2.53 & 1.60 & 2.69 & 2.58 \\
                                      & Ours        & \textbf{2.29} & \textbf{2.61} & \textbf{2.45} & \textbf{2.40} & \textbf{2.83} & \textbf{2.46} & \textbf{1.39} & \textbf{2.63} & \textbf{2.43} \\
\bottomrule 
\bottomrule
\end{tabular}
\caption{\reviewadd{The accuracy, peak signal-to-noise ratio (PSNR), structural similarity (SSIM), and binary cross entropy metrics for the simulation experiment in Section \ref{sec:simulation}. The vision-only baseline uses the SegFormer+FastSAM network described in Section \ref{sec:implementation}. Our proposed approach uses the same network for visual classification and uses simulated property measurements and Algorithm \ref{alg:moments} to update semantic predictions. Property measurements are shown to improve semantic prediction for all material classes across all metrics.}}
\label{table:metrics}
\end{table*}

\subsection{Habitat Simulation Evaluation} \label{subsec:habitat_eval}
We perform additional experiments in the Habitat simulator \cite{savva2019habitat} which provides ground-truth RGB, depth, and semantic data for various indoor scenes.
We demonstrate the performance of the Selmap \cite{ewen2022these} baseline using three pre-trained semantic segmentation models: the Dense Material Segmentation Dataset model \cite{upchurch2022materials}, the BERT image transformer \cite{bao2021bert}, and OneFormer \cite{jain2022oneformer}.
We note that the Selmap baseline is itself not a semantic segmentation model, but rather a method for probabilistically fusing semantic segmentation images from these networks within a 3D map.

Incorrectly classified regions are chosen by comparing the predicted semantic classifications with the ground-truth semantic information from the Habitat simulator.
We simulate friction measurements for incorrectly classified regions by randomly sampling from the Gaussian distribution whose parameters are specified in Table \ref{table:friction} using the corresponding ground-truth semantic label provided by the simulator.
Algorithm \ref{alg:moments} is then invoked to compute the posterior predictive likelihood of the semantic label.

We perform these simulation experiments for 5 distinct scenes in the Habitat simulator and use 10 RGB-D images per scene with known pose. 
Shown in Figure \ref{fig:replica_demo} are the results for Selmap alongside our proposed approach and ground-truth semantic labels for a sample scene from the Habitat simulator.

Table \ref{table:metrics} contains the evaluation metrics, including peak signal-to-noise ratio (PSNR), structural similarity (SSIM), binary cross entropy, and mean pixel-wise accuracy, used to evaluate our proposed approach against the vision-only Selmap \cite{ewen2022these} baseline.
These metrics are computed using both the Dense Material Segmentation Dataset results in Section \ref{subsec:dmsd_eval} and the Habitat simulation results discussed above.
We note that by leveraging tactile sensing, all evaluation metrics improve when compared to the vision-only baseline.

This experiment highlights that by conditioning semantic classifications on physical properties, one improve the accuracy of semantic classifications.

\section{Hardware Experiments} \label{sec:experiments}
Next, we validate our method on several hardware demonstrations and compare against existing semantic mapping and property estimation approaches.
In the first demonstration, we validate the results from simulation and show that our approach is capable of correcting erroneous semantic classifications by measuring the coefficient of friction.
We then provide two case studies for the utility of estimating physical properties.
For the first case study, our proposed approach is used within a risk-aware legged locomotion planner to update the locomotion gait of a quadruped when hazardous terrain is perceived.
In the second case study, we demonstrate our approach is able to estimate affordance-based properties in a door-opening scenario.
Both case studies are presented on the accompanying project webpage.

\subsection{Semantic Prediction Using Property Measurements} \label{subsec:first_exp}

The end-effector of a Kinova Gen 3 robotic arm and a Spot Arm are used to collect friction measurements using built-in wrist-mounted force-torque sensors.
For this experiment, 5 frames from the RGB-D stream are used to initialize the semantic TSDF map.
Each RGB-D image is semantically segmented using the trained network and projected into the global coordinate frame using the aligned depth image.
The recursive vision-based semantic update described in Section \ref{sec:vision} is applied for each semantic point cloud.
This initializes the $\boldsymbol{\alpha}$ parameters in \eqref{eq:property_comp} used to compute the initial semantic classification weights for the measurement likelihood.
The Gaussian parameters are initialized via Table \ref{table:friction} and the remaining Dirichlet Normal-Inverse-Gamma parameters are initialized as per Section \ref{sec:simulation}.

\begin{figure*}
    \centering
    \includegraphics[width=2\columnwidth]{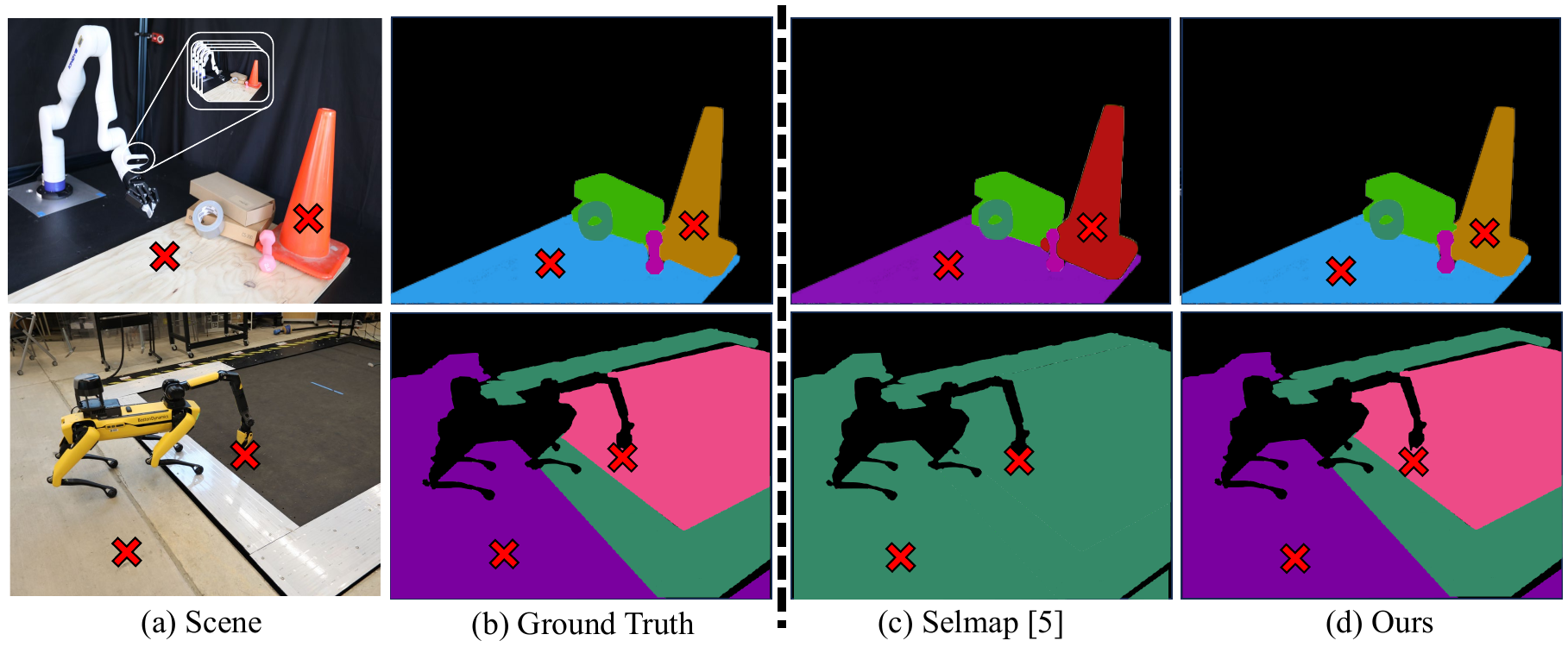}
     \caption{A semantic segmentation task is shown and the proposed method is compared against the semantic mapping approach from \cite{ewen2022these} which is called Selmap. The expected semantic class is shown. 
     a) The input image with measurement locations shown using an X and associated b) ground truth semantic labels are provided.
     Both Selmap and our approach use the same SegFormer + FastSAM pre-trained semantic segmentation network specified in Section \ref{sec:implementation}. 
     This network incorrectly predicts the semantic labels for several regions within the scene.
     c) This incorrect labelling is reflected in the Selmap implementation as the visual-based semantic mapping approach is unable to correct these erroneous predictions. 
     By exploiting a tactile sensing modality, d) our approach is able to correct the erroneous semantic predictions and correctly predict the semantic labels of the objects within the scene.}
     \label{fig:kinova_demo}
\end{figure*}

We compare our approach to the vision-only recursive semantic mapping approach of \cite{ewen2022these} which probabilistically filters semantic segmentation images within a metric map.
The same SegFormer-FastSAM semantic segmentation network is used for both methods.
As shown in Figure \ref{fig:kinova_demo}, the network incorrectly classifies various objects in the scenes.
When only vision-based semantic classifications are considered \cite{ewen2022these}, the erroneous predictions from the network are unable to be corrected (Figure \ref{fig:kinova_demo}c).

Using the proposed method, a user specifies the location for friction measurements to be taken (Figure \ref{fig:kinova_demo}a).
Algorithm \ref{alg:moments} then computes the posterior semantic classification weights using the friction measurements as input.
This corrects the expected semantic classification, as shown in Figure \ref{fig:kinova_demo}d.

We ran the experiment on two indoor scenes with two friction measurements each as depicted in Figure \ref{fig:kinova_demo}.
As shown in Figure \ref{fig:kinova_demo}, the semantic prediction accuracy increases using the proposed method and matches the ground-truth semantic labels.
In contrast, \cite{ewen2022these} is unable to correct the erroneous predictions from the semantic segmentation network, resulting in incorrect semantic predictions.
This demonstrates that the simulation results shown in Section \ref{sec:simulation} translate to hardware experiments.

\subsection{Friction Estimation Using Semantics} \label{subsec:second_exp}

We demonstrate the utility of our proposed property estimation method using a case study involving a challenging legged locomotion traversal task of crossing an icy surface.
We implement a locomotion planner similar to \cite{filitchkin2012feature} that switches between predetermined static and dynamic gaits depending on the expected value for the coefficient of friction of the terrain to be traversed.
A static gait is used when the expected value for the coefficient of friction falls below a threshold of $\E[\psi] \leq 0.25$.

A stream of RGB-D images is used to build a semantic TSDF map of the robot's environment as outlined in Section \ref{sec:vision}.
The resultant semantic TSDF map predicts that the region the robot must traverse may have a low coefficient of friction and a measurement is therefore taken to better approximate the value.

The Dirichlet Normal-Inverse-Gamma prior parameters are initialized as described in Section \ref{sec:simulation}.
We use the manipulator mounted to the legged robot to take friction measurements of the ground in front of the robot using a built-in wrist-mounted force-torque sensor via \eqref{eq:friction}.
Algorithm \ref{alg:moments} then computes the posterior estimate for the coefficient of friction (Figure \ref{fig:main}).

\begin{figure} [!t]
    \centering
    \includegraphics[width=\columnwidth]{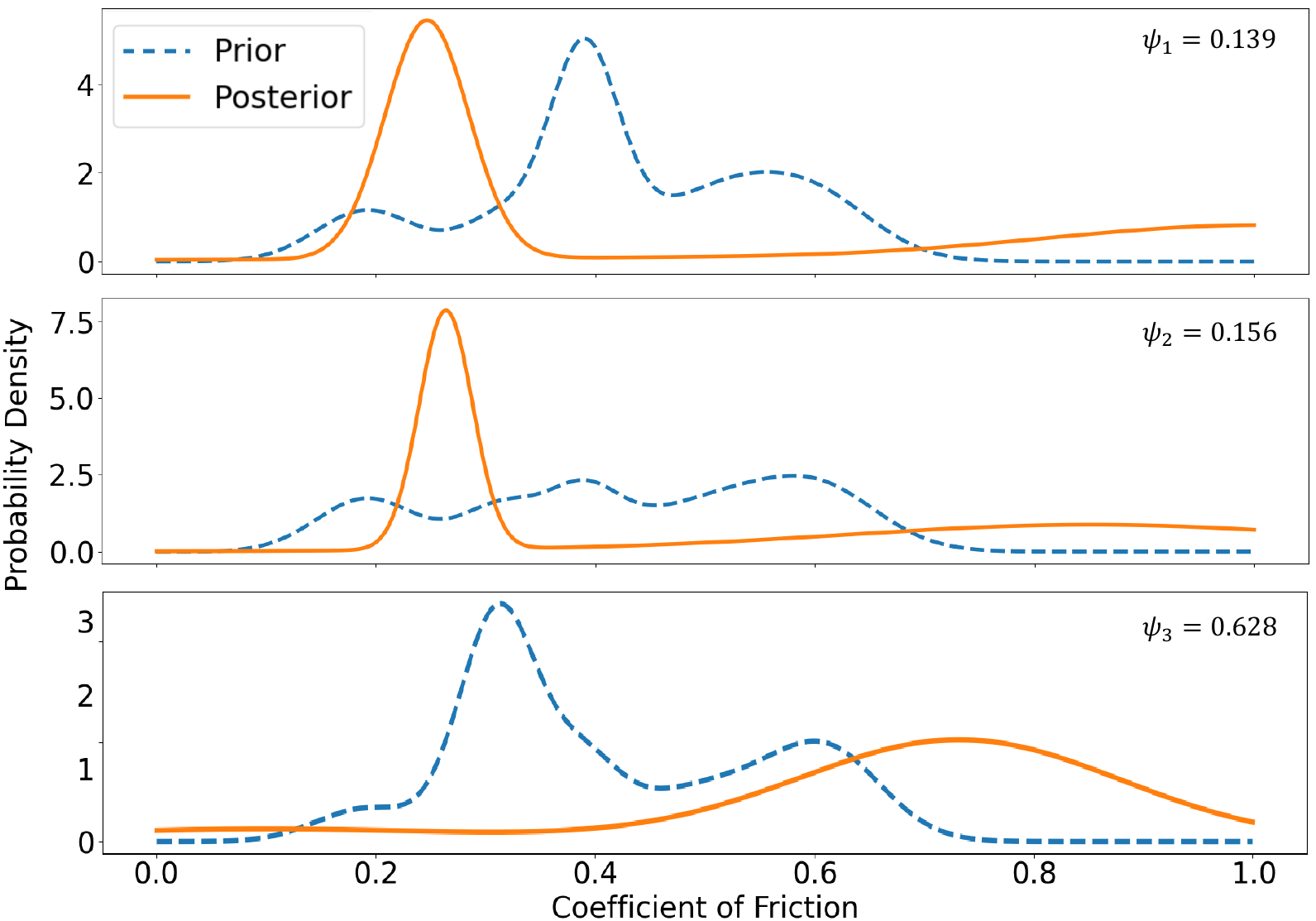}
    \caption{The prior (dotted) and posterior (solid) friction estimates for the three ice traversal demonstration runs. When friction measurements are taken of the ice before traversal, Algorithm \ref{alg:moments} predicts the posterior friction estimates. The friction measurements are $\psi_1=0.139$, $\psi_2=0.156$, and $\psi_3=0.628$ for trials 1, 2, and 3, respectively.}
    \label{fig:ice_demo_friction}
\end{figure}

The experiment was conducted six times in the same location during the same afternoon, three where the robot does not measure the coefficient of friction and three where it does.
The prior and posterior friction estimates for these six trials is shown in Figure \ref{fig:ice_demo_friction}.
The Selmap baseline incorrectly predicts the semantic label of the terrain for each run and mis-classifies the ice as either concrete or snow.
Since a friction measurement is not taken, the friction threshold $\E[\psi] \leq 0.25$ is not met due to mislabeling and the robot attempts to cross the ice using a dynamic gait, resulting in it slipping and falling for all three of these trials.
This comparison can be found at the project page website\footnote{\href{https://roahmlab.github.io/multimodal_mapping/}{https://roahmlab.github.io/multimodal\_mapping/}}.

For the subsequent two trials, friction measurements of $\psi_1=0.139$ and $\psi_2=0.156$ are taken for each trial, respectively.
After computing the posterior friction estimates via Algorithm \ref{alg:moments}, the friction threshold is met and the robot switches to a static, stable gait to cross the ice.
This results in successful ice traversal for these two trials.
In the following third trial, a friction measurement of $\psi_3=0.628$ is taken due to erroneous end-effector placement.
The computed posterior is shown in Figure \ref{fig:ice_demo_friction} and shows a coefficient of friction with a large expected value which exceeds the threshold.
For this trial, the robot remains in a dynamic gait and slips on the ice during traversal.
A video of a successful ice traversal trial is provided on the accompanying project webpage.

\begin{figure} [!t]
    \centering
    \includegraphics[width=\columnwidth]{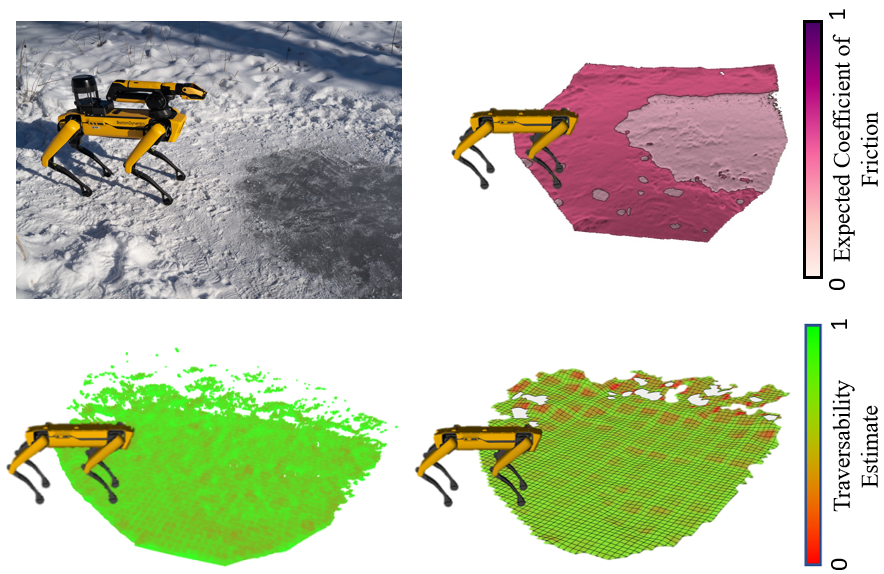}
    \caption{The performance of the proposed approach after a friction measurement is taken in comparison to the traversability estimate computed by \cite{agishev2022trav} (bottom left) and \cite{gan2022multitask} (bottom right). The proposed method is able to correctly predict that the surface in front of the robot has a low coefficient of friction which informs the locomotion planner to switch to a stable gait. In contrast, the traversability estimates predict the ice is traversable even when the robot is using a dynamic gait, which results in the robot slipping and falling.}
    \label{fig:trav_comp}
\end{figure}

We further compare our property estimation approach with two state-of-the-art traversability estimation methods designed for legged locomotion.
Both methods consider both geometric and semantic features and use a pre-trained neural network to estimate a traversability score \cite{agishev2022trav, gan2022multitask}.
Both traversability estimation methods predict that the icy surface is traversable as shown in Figure \ref{fig:trav_comp};
however, attempting to traverse this surface using a dynamic gait resulted in the robot slipping and falling.

Our proposed method is able to verify the property predictions estimated using visual information via tactile sensing and Bayesian inference whereas other state-of-the-art property estimation methods are unable to incorporate multiple sensing modalities.
This task highlights the utility of estimating physical properties such as the coefficient of friction to accomplish challenging tasks such as legged locomotion traversal across an icy surface.
State-of-the-art traversability estimation methods depend only on the terrain geometry and classification and thus do not provide the means by which a robot may adapt its traversal strategy to account for the properties (i.e. friction) which dictate traversability.

\subsection{Encoding Affordance-Based Properties}

Lastly, we present a scenario in which our approach is used to estimate affordance-based properties.
An affordance is a property of an object that defines its possible used \cite{gibson1977theory}.
In this case, we focus on the affordance of a door handle, which affords either a push or pull action, as the semantic classification and examine the force required to open the door as the property of interest.

\begin{figure} [!t]
    \centering
    \includegraphics[width=\columnwidth]{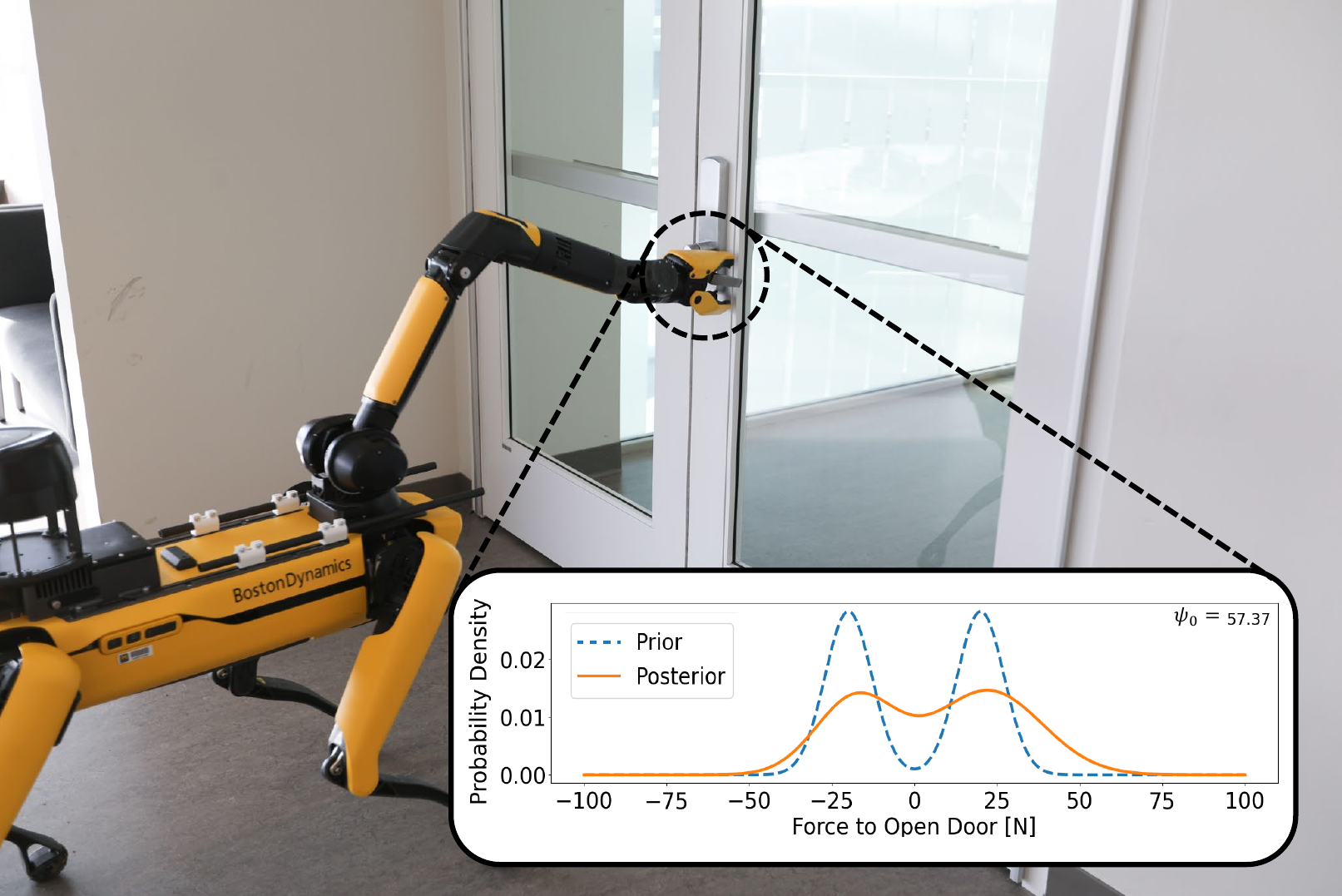}
    \caption{The proposed method is capable of encoding affordance-based properties, in this case the force required to open a door. 
    This door opening force is initialized as a multi-modal distribution with pushing and pulling being equally likely with a magnitude of 20N.
    The robot applies a pulling force of 57N on the door, resulting in it swinging open.
    The posterior estimate for the force required to open a door is computed and shown in the embedded plot.}
    \label{fig:door_opening}
\end{figure}

As shown in Figure \ref{fig:door_opening}, a mobile manipulator is used to measure the force required to open a door.
Negative forces constitute pushing while positive forces constitute pulling.
We equally weight each push/pull affordance and initialize a mean force magnitude of $20N$ and a variance of $10N$.

To measure the force required to open the door, the robot begins applying a pulling force $10N$ and ramps this up to $70N$ over a period of 3 seconds.
At $57N$ the door begins to open and this value is used as the property measurement for Algorithm \ref{alg:moments}.
The door opening force posterior is show in Figure \ref{fig:door_opening} in the lower right-hand corner.
After invoking Algorithm \ref{alg:moments} the push/pull affordances are still almost equally weighed.
Additionally, the variance for the pulling force mode has increased, mirroring the fact that a higher-than-expected force was required to open the door.
This door opening experiment is demonstrated on the accompanying project webpage.

\begin{figure} [!t]
    \centering
    \includegraphics[width=\columnwidth]{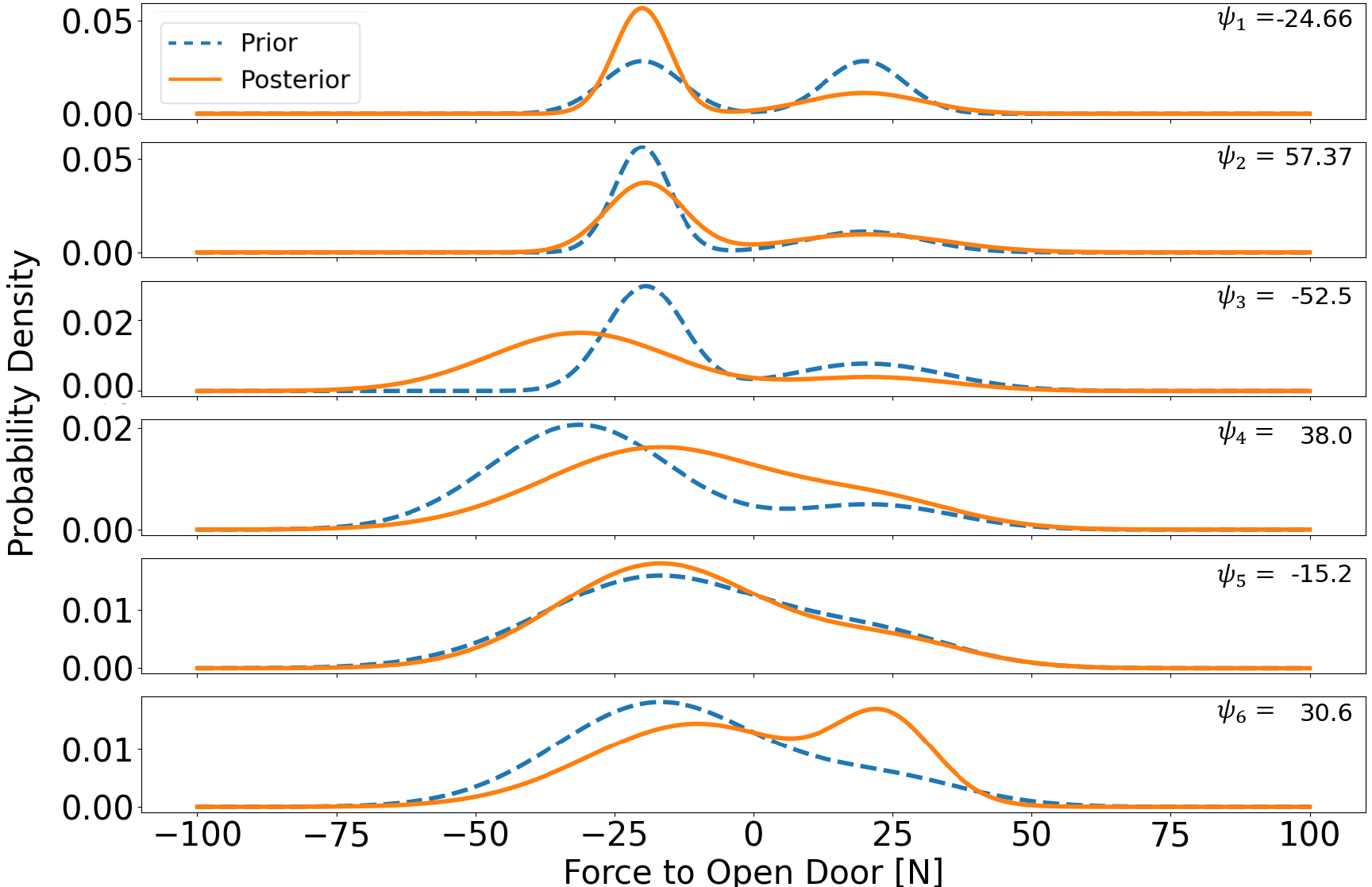}
    \caption{The prior and posterior door opening force estimates running Algorithm \ref{alg:moments} sequentially with six experiments, meaning the posterior for the previous experiment is the prior for the current experiment.
    The force required to open the door is indicated in the upper right-hand corner for each experiment. 
    After measuring the force required to open 6 doors, the posterior estimate has a high variance with an expected value near $0N$, reflecting the equal likelihood that a door is either push or pull to open.}
    \label{fig:many_doors}
\end{figure}

Next, we test the force required to open six different doors, three pull-to-open and three push-to-open, and use Algorithm \ref{alg:moments} to compute the posterior door opening force estimate sequentially after each trail.
Figure \ref{fig:many_doors} shows the prior and posterior opening force estimate after each trial.
After six trials the estimated door opening force posterior has maintained two modes corresponding to the push or pull force modalities, however the variance has increased, representing the varied door opening forces seen in the experiments compared to the initial estimate.

This case study demonstrates the utility of the proposed method for affordance-based property estimation and opens an avenue for future research given recent advances in affordance-based planning \cite{rezapour2019towards}.

\section{Discussion and Conclusion} \label{sec:conclusion}
We propose a method for jointly estimating semantic labels and physical properties.
We model semantic classifications and physical properties probabilistically enabling closed-form Bayesian inference given visual semantic predictions and tactile property measurements.
We demonstrate that by leveraging this multi-modal probabilistic approach we outperform the vision-only baselines across all evaluation metrics in simulation and provide hardware experiments demonstrating the efficacy of the proposed approach.
Additionally, we provide two case studies which motivate the utility of physical property estimation demonstrated in the proposed method.

One drawback of our approach is that the accuracy of our method is reliant on the accuracy of the semantic classification network.
If a semantic classification network consistently predicts incorrect semantic labels over a long duration, the initial Dirichlet parameters will skew heavily towards the incorrect semantic label and be difficult to correct with property measurements.
Additionally, like many voxel-based semantic mapping frameworks, our method is constrained by memory limitations which restrict the size of the map.
We aim to address these concerns in future work.

Tactile and visual sensing modalities were used for the proposed method, however additional sensing modalities may also be used in place of or in addition to tactile sensing.
This is an avenue for future work, \reviewadd{along with incorporating tactile sensing noise}.
Motivated by recent work in legged locomotion \cite{miki2022learning}, we aim to use the property estimates computed using our method as inputs to locomotion controllers.
Such a supervisory signal would enable robots to change their gait before transitioning between surfaces and potentially enable safer, robust locomotion policies.
A similar use for property uncertainty may be found in assessing locomotion risk, and a risk threshold may be used to determine when new tactile or visual measurements are required.

Lastly, our presented method provides a link between visual and tactile sensing modalities through physical property estimates and future work will aim to exploit this relationship for active perception.
Current active perception methods are primarily concerned with geometric reconstruction, however as demonstrated in Section \ref{sec:experiments}, physical properties play an important role in robotic task completion.
As such, future work will explore how to leverage uncertainty in semantic and property estimates to find where in the environment a robot should explore, or where to take tactile measurements to decrease semantic uncertainty computed from visual data.

\renewcommand{\bibfont}{\normalfont\small}
{\renewcommand{\markboth}[2]{}% Remove header adjustment
\printbibliography}

@article{guo2018review,
  title={A review of semantic segmentation using deep neural networks},
  author={Guo, Yanming and Liu, Yu and Georgiou, Theodoros and Lew, Michael S},
  journal={International journal of multimedia information retrieval},
  volume={7},
  pages={87--93},
  year={2018},
  publisher={Springer}
}

@article{hao2020brief,
  title={A brief survey on semantic segmentation with deep learning},
  author={Hao, Shijie and Zhou, Yuan and Guo, Yanrong},
  journal={Neurocomputing},
  volume={406},
  pages={302--321},
  year={2020},
  publisher={Elsevier}
}

@inproceedings{strudel2021segmenter,
  title={Segmenter: Transformer for semantic segmentation},
  author={Strudel, Robin and Garcia, Ricardo and Laptev, Ivan and Schmid, Cordelia},
  booktitle={Proceedings of the IEEE/CVF international conference on computer vision},
  pages={7262--7272},
  year={2021}
}

@article{garcia2018,
  title={A survey on deep learning techniques for image and video semantic segmentation},
  author={Garcia-Garcia, Alberto and Orts-Escolano, Sergio and Oprea, Sergiu and Villena-Martinez, Victor and Martinez-Gonzalez, Pablo and Garcia-Rodriguez, Jose},
  journal={Applied Soft Computing},
  volume={70},
  pages={41--65},
  year={2018},
  publisher={Elsevier}
}

@inproceedings{lin2014microsoft,
  title={Microsoft coco: Common objects in context},
  author={Lin, Tsung-Yi and Maire, Michael and Belongie, Serge and Hays, James and Perona, Pietro and Ramanan, Deva and Doll{\'a}r, Piotr and Zitnick, C Lawrence},
  booktitle={Computer Vision--ECCV 2014: 13th European Conference, Zurich, Switzerland, September 6-12, 2014, Proceedings, Part V 13},
  pages={740--755},
  year={2014},
  organization={Springer}
}

@article{xiang2017rnn,
  title={{DA-RNN}: Semantic mapping with data associated recurrent neural networks},
  author={Xiang, Yu and Fox, Dieter},
  journal={arXiv preprint arXiv:1703.03098},
  year={2017}
}

@inproceedings{deng2009imagenet,
  title={Imagenet: A large-scale hierarchical image database},
  author={Deng, Jia and Dong, Wei and Socher, Richard and Li, Li-Jia and Li, Kai and Fei-Fei, Li},
  booktitle={2009 IEEE conference on computer vision and pattern recognition},
  pages={248--255},
  year={2009},
  organization={Ieee}
}

@article{kuipers1991robot,
  title={A robot exploration and mapping strategy based on a semantic hierarchy of spatial representations},
  author={Kuipers, Benjamin and Byun, Yung-Tai},
  journal={Robotics and autonomous systems},
  volume={8},
  number={1-2},
  pages={47--63},
  year={1991},
  publisher={Elsevier}
}

@inproceedings{upchurch2022materials,
title = {A Dense Material Segmentation Dataset for Indoor and Outdoor Scene Parsing},
booktitle = {ECCV},
author = {Paul Upchurch* and Ransen Niu*},
year = {2022},
URL = {https://arxiv.org/abs/2207.10614}
}

@book{wang2019road,
  title={Road terrain classification technology for autonomous vehicle},
  author={Wang, Shifeng},
  year={2019},
  publisher={Springer}
}

@inproceedings{noh2021surface,
  title={Surface material dataset for robotics applications ({SMDRA}): A dataset with friction coefficient and {RGB-D} for surface segmentation},
  author={Noh, Donghun and Nam, Hyunwoo and Ahn, Min Sung and Chae, Hosik and Lee, Sangjoon and Gillespie, Kyle and Hong, Dennis},
  booktitle={2020 25th International Conference on Pattern Recognition (ICPR)},
  pages={6275--6281},
  year={2021},
  organization={IEEE}
}

@inproceedings{brandao2016friction,
  title={Friction from vision: A study of algorithmic and human performance with consequences for robot perception and teleoperation},
  author={Brandao, Martim and Hashimoto, Kenji and Takanishi, Atsuo},
  booktitle={2016 IEEE-RAS 16th International Conference on Humanoid Robots (Humanoids)},
  pages={428--435},
  year={2016},
  organization={IEEE}
}

@article{kostavelis2015semantic,
  title={Semantic mapping for mobile robotics tasks: A survey},
  author={Kostavelis, Ioannis and Gasteratos, Antonios},
  journal={Robotics and Autonomous Systems},
  volume={66},
  pages={86--103},
  year={2015},
  publisher={Elsevier}
}

@inproceedings{mccormac2017semanticfusion,
  title={Semanticfusion: Dense 3d semantic mapping with convolutional neural networks},
  author={McCormac, John and Handa, Ankur and Davison, Andrew and Leutenegger, Stefan},
  booktitle={2017 IEEE International Conference on Robotics and automation (ICRA)},
  pages={4628--4635},
  year={2017},
  organization={IEEE}
}

@inproceedings{sunderhauf2017meaningful,
  title={Meaningful maps with object-oriented semantic mapping},
  author={S{\"u}nderhauf, Niko and Pham, Trung T and Latif, Yasir and Milford, Michael and Reid, Ian},
  booktitle={2017 IEEE/RSJ International Conference on Intelligent Robots and Systems (IROS)},
  pages={5079--5085},
  year={2017},
  organization={IEEE}
}

@article{ewen2022these,
  title={These Maps are Made for Walking: Real-Time Terrain Property Estimation for Mobile Robots},
  author={Ewen, Parker and Li, Adam and Chen, Yuxin and Hong, Steven and Vasudevan, Ram},
  journal={IEEE Robotics and Automation Letters},
  volume={7},
  number={3},
  pages={7083--7090},
  year={2022},
  publisher={IEEE}
}

@article{gan2022multitask,
  title={Multitask learning for scalable and dense multilayer {B}ayesian map inference},
  author={Gan, Lu and Kim, Youngji and Grizzle, Jessy W and Walls, Jeffrey M and Kim, Ayoung and Eustice, Ryan M and Ghaffari, Maani},
  journal={IEEE Transactions on Robotics},
  volume={39},
  number={1},
  pages={699--717},
  year={2022},
  publisher={IEEE}
}

@article{wilson2023convbki,
  title={Conv{BKI}: Real-Time Probabilistic Semantic Mapping Network with Quantifiable Uncertainty},
  author={Wilson, Joey and Fu, Yuewei and Friesen, Joshua and Ewen, Parker and Capodieci, Andrew and Jayakumar, Paramsothy and Barton, Kira and Ghaffari, Maani},
  journal={arXiv preprint arXiv:2310.16020},
  year={2023}
}

@article{couprie2013indoor,
  title={Indoor semantic segmentation using depth information},
  author={Couprie, Camille and Farabet, Cl{\'e}ment and Najman, Laurent and LeCun, Yann},
  journal={arXiv preprint arXiv:1301.3572},
  year={2013}
}

@inproceedings{wang2015towards,
  title={Towards unified depth and semantic prediction from a single image},
  author={Wang, Peng and Shen, Xiaohui and Lin, Zhe and Cohen, Scott and Price, Brian and Yuille, Alan L},
  booktitle={Proceedings of the IEEE conference on computer vision and pattern recognition},
  pages={2800--2809},
  year={2015}
}

@article{sun2020fuseseg,
  title={Fuse{S}eg: Semantic segmentation of urban scenes based on {RGB} and thermal data fusion},
  author={Sun, Yuxiang and Zuo, Weixun and Yun, Peng and Wang, Hengli and Liu, Ming},
  journal={IEEE Transactions on Automation Science and Engineering},
  volume={18},
  number={3},
  pages={1000--1011},
  year={2020},
  publisher={IEEE}
}

@inproceedings{silberman2012indoor,
  title={Indoor segmentation and support inference from {RGBD} images},
  author={Silberman, Nathan and Hoiem, Derek and Kohli, Pushmeet and Fergus, Rob},
  booktitle={Computer Vision--ECCV 2012: 12th European Conference on Computer Vision, Florence, Italy, October 7-13, 2012, Proceedings, Part V 12},
  pages={746--760},
  year={2012},
  organization={Springer}
}

@article{valada2020self,
  title={Self-supervised model adaptation for multimodal semantic segmentation},
  author={Valada, Abhinav and Mohan, Rohit and Burgard, Wolfram},
  journal={International Journal of Computer Vision},
  volume={128},
  number={5},
  year={2020},
  publisher={Springer}
}

@inproceedings{kim2018season,
  title={Season-invariant semantic segmentation with a deep multimodal network},
  author={Kim, Dong-Ki and Maturana, Daniel and Uenoyama, Masashi and Scherer, Sebastian},
  booktitle={Field and Service Robotics: Results of the 11th International Conference},
  year={2018},
  organization={Springer}
}

@article{gibson1977theory,
  title={The theory of affordances},
  author={Gibson, James J},
  journal={Hilldale, USA},
  volume={1},
  number={2},
  year={1977}
}

@article{papadakis2013terrain,
  title={Terrain traversability analysis methods for unmanned ground vehicles: A survey},
  author={Papadakis, Panagiotis},
  journal={Engineering Applications of Artificial Intelligence},
  volume={26},
  number={4},
  pages={1373--1385},
  year={2013},
  publisher={Elsevier}
}

@inproceedings{kim2006traversability,
  title={Traversability classification using unsupervised on-line visual learning for outdoor robot navigation},
  author={Kim, Dongshin and Sun, Jie and Oh, Sang Min and Rehg, James M and Bobick, Aaron F},
  booktitle={Proceedings 2006 IEEE International Conference on Robotics and Automation, 2006. ICRA 2006.},
  pages={518--525},
  year={2006},
  organization={IEEE}
}

@inproceedings{frey2022locomotion,
  title={Locomotion policy guided traversability learning using volumetric representations of complex environments},
  author={Frey, Jonas and Hoeller, David and Khattak, Shehryar and Hutter, Marco},
  booktitle={2022 IEEE/RSJ International Conference on Intelligent Robots and Systems (IROS)},
  pages={5722--5729},
  year={2022},
  organization={IEEE}
}

@article{frey2023fast,
  title={Fast Traversability Estimation for Wild Visual Navigation},
  author={Frey, Jonas and Mattamala, Matias and Chebrolu, Nived and Cadena, Cesar and Fallon, Maurice and Hutter, Marco},
  journal={arXiv preprint arXiv:2305.08510},
  year={2023}
}

@article{sferrazza2023power,
  title={The Power of the Senses: Generalizable Manipulation from Vision and Touch through Masked Multimodal Learning},
  author={Sferrazza, Carmelo and Seo, Younggyo and Liu, Hao and Lee, Youngwoon and Abbeel, Pieter},
  journal={arXiv preprint arXiv:2311.00924},
  year={2023}
}

@article{margolis2023learning,
  title={Learning to See Physical Properties with Active Sensing Motor Policies},
  author={Margolis, Gabriel B and Fu, Xiang and Ji, Yandong and Agrawal, Pulkit},
  journal={arXiv preprint arXiv:2311.01405},
  year={2023}
}

@article{cai2023evora,
  title={EVORA: Deep Evidential Traversability Learning for Risk-Aware Off-Road Autonomy},
  author={Cai, Xiaoyi and Ancha, Siddharth and Sharma, Lakshay and Osteen, Philip R and Bucher, Bernadette and Phillips, Stephen and Wang, Jiuguang and Everett, Michael and Roy, Nicholas and How, Jonathan P},
  journal={arXiv preprint arXiv:2311.06234},
  year={2023}
}

@inproceedings{izadi2011kinectfusion,
  title={Kinectfusion: real-time 3d reconstruction and interaction using a moving depth camera},
  author={Izadi, Shahram and Kim, David and Hilliges, Otmar and Molyneaux, David and Newcombe, Richard and Kohli, Pushmeet and Shotton, Jamie and Hodges, Steve and Freeman, Dustin and Davison, Andrew and others},
  booktitle={Proceedings of the 24th annual ACM symposium on User interface software and technology},
  pages={559--568},
  year={2011}
}

@incollection{nixon2012,
title = {Feature Extraction \& Image Processing for Computer Vision (Third edition)},
editor = {Mark S. Nixon and Alberto S. Aguado},
booktitle = {Feature Extraction \& Image Processing for Computer Vision (Third Edition)},
publisher = {Academic Press},
edition = {Third Edition},
address = {Oxford},
pages = {489-518},
year = {2012},
author = {Mark S. Nixon and Alberto S. Aguado}
}

@article{tu2014,
  title={The {D}irichlet-{M}ultinomial and {D}irichlet-{C}ategorical models for {B}ayesian inference},
  author={Tu, Stephen},
  journal={Computer Science Division, UC Berkeley},
  volume={2},
  year={2014}
}

@article{nguyen2021,
author = {Nguyen, Tran and Verdoja, Francesco and Abu-Dakka, Fares and Kyrki, Ville},
year = {2021},
month = {02},
pages = {1-1},
title = {Probabilistic Surface Friction Estimation Based on Visual and Haptic Measurements},
volume = {PP},
journal = {IEEE Robotics and Automation Letters},
}

@article{jaini2016online,
  title={Online and distributed learning of Gaussian mixture models by Bayesian moment matching},
  author={Jaini, Priyank and Poupart, Pascal},
  journal={arXiv preprint arXiv:1609.05881},
  year={2016}
}

@article{vaswani2017attention,
  title={Attention is all you need},
  author={Vaswani, Ashish and Shazeer, Noam and Parmar, Niki and Uszkoreit, Jakob and Jones, Llion and Gomez, Aidan N and Kaiser, Lukasz and Polosukhin, Illia},
  journal={Advances in neural information processing systems},
  volume={30},
  year={2017}
}

@article{xie2021segformer,
  title={Seg{F}ormer: Simple and efficient design for semantic segmentation with transformers},
  author={Xie, Enze and Wang, Wenhai and Yu, Zhiding and Anandkumar, Anima and Alvarez, Jose M and Luo, Ping},
  journal={Advances in Neural Information Processing Systems},
  volume={34},
  pages={12077--12090},
  year={2021}
}

@misc{chen2023semantic,
    title = {Semantic Segment Anything},
    author = {Chen, Jiaqi and Yang, Zeyu and Zhang, Li},
    howpublished = {\url{https://github.com/fudan-zvg/Semantic-Segment-Anything}},
    year = {2023}
}

@article{diaconis1979conjugate,
  title={Conjugate priors for exponential families},
  author={Diaconis, Persi and Ylvisaker, Donald},
  journal={The Annals of statistics},
  pages={269--281},
  year={1979},
  publisher={JSTOR}
}

@misc{minka2000estimating,
  title={Estimating a {D}irichlet distribution},
  author={Minka, Thomas},
  year={2000},
  publisher={Technical report, MIT}
}

@book{bernardo2009bayesian,
  title={Bayesian theory},
  author={Bernardo, Jos{\'e} M and Smith, Adrian FM},
  volume={405},
  year={2009},
  publisher={John Wiley \& Sons}
}

@inproceedings{filitchkin2012feature,
  title={Feature-based terrain classification for littledog},
  author={Filitchkin, Paul and Byl, Katie},
  booktitle={2012 IEEE/RSJ International Conference on Intelligent Robots and Systems},
  pages={1387--1392},
  year={2012},
  organization={IEEE}
}

@book{anderson2012optimal,
  title={Optimal filtering},
  author={Anderson, Brian DO and Moore, John B},
  year={2012},
  publisher={Courier Corporation}
}

@article{thrun2002probabilistic,
  title={Probabilistic robotics},
  author={Thrun, Sebastian},
  journal={Communications of the ACM},
  volume={45},
  number={3},
  pages={52--57},
  year={2002},
  publisher={ACM New York, NY, USA}
}

@ARTICLE{agishev2022trav,
  author={Agishev, Ruslan and Petříček, Tomáš and Zimmermann, Karel},
  journal={IEEE Robotics and Automation Letters},
  title={Trajectory Optimization Using Learned Robot-Terrain Interaction Model in Exploration of Large Subterranean Environments},
  year={2022},
  volume={7},
  number={2},
  pages={3365-3371},
  doi={10.1109/LRA.2022.3147332}
}

@article{rezapour2019towards,
  title={Towards affordance detection for robot manipulation using affordance for parts and parts for affordance},
  author={Rezapour Lakani, Safoura and Rodr{\'\i}guez-S{\'a}nchez, Antonio J and Piater, Justus},
  journal={Autonomous Robots},
  volume={43},
  pages={1155--1172},
  year={2019},
  publisher={Springer}
}

@article{miki2022learning,
  title={Learning robust perceptive locomotion for quadrupedal robots in the wild},
  author={Miki, Takahiro and Lee, Joonho and Hwangbo, Jemin and Wellhausen, Lorenz and Koltun, Vladlen and Hutter, Marco},
  journal={Science Robotics},
  volume={7},
  number={62},
  year={2022},
  publisher={American Association for the Advancement of Science}
}

@inproceedings{savva2019habitat,
  title={Habitat: A platform for embodied ai research},
  author={Savva, Manolis and Kadian, Abhishek and Maksymets, Oleksandr and Zhao, Yili and Wijmans, Erik and Jain, Bhavana and Straub, Julian and Liu, Jia and Koltun, Vladlen and Malik, Jitendra and others},
  booktitle={Proceedings of the IEEE/CVF international conference on computer vision},
  pages={9339--9347},
  year={2019}
}

@inproceedings{bao2021bert,
  author= {Hangbo Bao and Li Dong and Furu Wei},
  title= {BEiT: {BERT} Pre-Training of Image Transformers},
  journal   = {CoRR},
  year      = {2021},
  bibsource = {dblp computer science bibliography, https://dblp.org}
}

@article{jain2022oneformer,
      title={{OneFormer: One Transformer to Rule Universal Image Segmentation}},
      author={Jitesh Jain and Jiachen Li and MangTik Chiu and Ali Hassani and Nikita Orlov and Humphrey Shi},
      journal={arXiv}, 
      year={2022}
    }

@inproceedings{shneier2006performance,
  title={Performance evaluation of a terrain traversability learning algorithm in the {DARPA LAGR} program},
  author={Shneier, Michael and Shackleford, Will and Hong, Tsai and Chang, Tommy},
  booktitle={Performance Metrics for Intelligent Systems Workshop, Gaithersburg, MD, USA},
  pages={103--110},
  year={2006}
}

@ARTICLE{hornung13auro,
  author = {Armin Hornung and Kai M. Wurm and Maren Bennewitz and Cyrill
  Stachniss and Wolfram Burgard},
  title = {{OctoMap}: An Efficient Probabilistic {3D} Mapping Framework Based
  on Octrees},
  journal = {Autonomous Robots},
  year = 2013,
  doi = {10.1007/s10514-012-9321-0},
  note = {Software available at \url{https://octomap.github.io}}
}

\appendices
\section{Proof of Theorem \ref{thm:dir_cat_conj}} \label{app:A}

Let $p(\boldsymbol{\theta} | \boldsymbol{\alpha})$ be the prior belonging to the family of Dirichlet distributions and $p(z | \boldsymbol{\theta})$ the likelihood belonging to the family of Categorical distributions.
Let $\Z = \{z_1, \dots, z_n\}$ be a set of measurements drawn from the Categorical likelihood.

The Law of Total Probability is used to compute the posterior distribution:
\begin{align}
    p(\boldsymbol{\theta} | \Z, \boldsymbol{\alpha}) & \propto p(\boldsymbol{\theta}, \Z, \boldsymbol{\alpha}) \\
    &= p(\boldsymbol{\theta} | \boldsymbol{\alpha}) \prod_{z_i \in \Z} p(z_i | \boldsymbol{\theta})  \label{eq:dir_cat_conj_proof},
\end{align}
where the equality is due to conditional independence.

Then, substituting \eqref{eq:categorical} and \eqref{eq:dirichlet} into \eqref{eq:dir_cat_conj_proof}, we arrive at:

\begin{align}
    p(\boldsymbol{\theta} | \boldsymbol{\alpha}) \prod_{z_i \in \Z} p(z_i | \boldsymbol{\theta}) &\propto \prod_{j=1}^k \theta_j^{\alpha_j-1} \prod_{z_i \in \Z} \prod_{j=1}^k \theta_j^{1\{z_i = j\}} \\
    &= \prod_{j=1}^k \theta_j^{\alpha_j-1 + \sum_{z_i \in \Z} 1\{z_i = j\}}.
\end{align}

This is exactly the Dirichlet distribution with parameters

\begin{equation}
    \tilde{\alpha}_j = \alpha_j + \sum_{z_i \in \mathcal{Z}} 1\{z_i = j\}.
\end{equation}

\section{Proof of Theorem \ref{thm:bmm_conj}} \label{app:B}

Let $p(\Theta | \Psi)$ be the prior belonging to the family of Dirichlet Normal-Inverse-Gamma distributions and $p(\psi | \Theta)$ the likelihood belonging to the family of Gaussian mixtures.
Let $\psi$ be a measurements drawn from the Gaussian mixture likelihood.

From Theorem \ref{thm:true_posterior}, the posterior of $\Theta$ is given by:

\begin{align} \label{eq:conj_post_proof}
    p(\Theta | \psi, \Psi) = \frac{1}{M}\sum_{j=1}^k & c_j Dir(\boldsymbol{w} | \tilde{\boldsymbol{a}}_j) \cdot \nonumber \\
    &\cdot \NIG(\mu_j, \sigma^2_j | \Tilde{\tau}_j, \Tilde{\kappa}_j, \Tilde{\beta}_j, \Tilde{\gamma}_j) \cdot \nonumber\\
    &\cdot \prod_{i \neq j}^k \NIG(\mu_i, \sigma^2_i | \tau_i, \kappa_i, \beta_i, \gamma_i),
\end{align}
with parameters computed as per \eqref{eq:true_posterior}-\eqref{eq:true_posterior_end}. 

It is evident that the number of terms in this posterior grows exponentially for each measurement.
To overcome this challenge, we project \eqref{eq:conj_post_proof} onto the family of Dirichlet Normal-Inverse-Gamma product distributions using the method of moments.
We now prove the set of sufficient moments for the Dirichlet Normal-Inverse-Gamma product distribution is $\mathbb{S} = \{\mu_i, \sigma^2_i, \sigma_i^4, \mu_i^2 \sigma^2_i, w_i, w_i^2\}_{i=1}^k$.

From \eqref{eq:bmm}, it is evident that $\boldsymbol{w}$ drawn from the Dirichlet distribution and $\Phi_i = \{\mu_i, \sigma_i\}$ drawn from the Normal-Inverse-Gamma distributions are independent random variables.
Therefore, to find the sufficient moments of the Dirichlet Normal-Inverse-Gamma product distribution, we may use the sufficient moments for the Dirichlet and Normal-Inverse-Gamma distributions.
The sufficient moments for the Dirichlet distribution are $\{\boldsymbol{w}, \boldsymbol{w}^2\}$ \cite{minka2000estimating} and the sufficient moments of the $i$-th Normal-Inverse-Gamma distribution are $\{\mu_i, \sigma^2_i, \sigma_i^4, \mu_i^2 \sigma^2_i\}$ {\cite[Chapter 4.5]{bernardo2009bayesian}}.
The concatenation of these sufficient moments then constitute the sufficient moments of \eqref{eq:bmm}.

The parameters of the Dirichlet Normal-Inverse-Gamma product distribution are then computed via \eqref{eq:true_posterior}-\eqref{eq:true_posterior_end} using the sufficient statistics as derived in \cite{minka2000estimating} and \cite{bernardo2009bayesian}, respectively.

\begin{figure} [t!]
    \centering
    \includegraphics[width=\columnwidth]{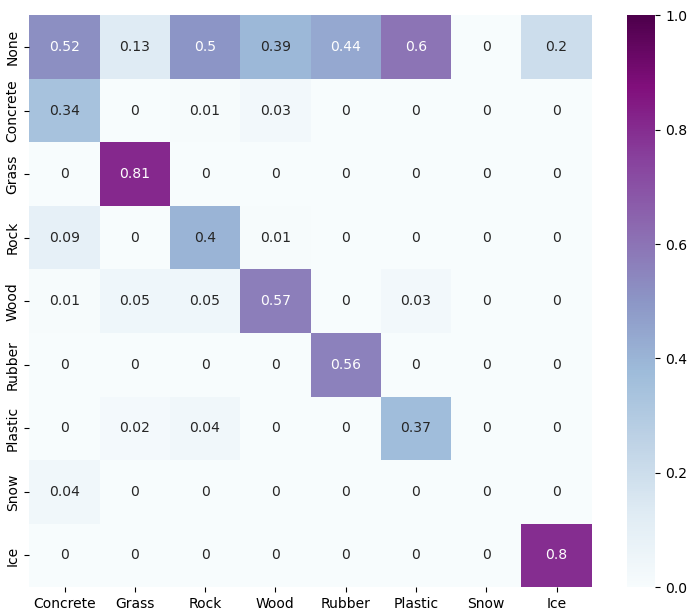}
    \caption{\reviewadd{Confusion matrix for the combined SegFormer+FastSAM semantic segmentation network computed using a randomly chosen subset of the test set of the Dense Material Segmentation Dataset \cite{upchurch2022materials}. There are no images of snow in this subset, as demonstrated by the confusion matrix.}}
    \label{fig:confusion_matrix}
\end{figure}

\subsection{Computing the Sufficient Statistics}

The sufficient statistics $\mathbb{S} = \{\mu_i, \sigma^2_i, \sigma_i^4, \mu_i^2 \sigma^2_i, w_i, w_i^2\}_{i=1}^k$ for \eqref{eq:bmm} must be computed from \eqref{eq:conj_post_proof}.
It is possible to use numerical integration to compute the sufficient moments via \eqref{eq:moments}, however we demonstrate a more computationally efficient method here by exploiting the linearity of the expectation operator and the independence of variables within \eqref{eq:conj_post_proof}.

We derive the first sufficient moment $\E[\mu_i]$ and leave the remaining moment derivations to the reader.
We start by noting the independence of $\boldsymbol{w}$ and each set of parameters $\Phi_i = \{\mu_i, \sigma^2_i\}$.
Thus, to compute $\E[\mu_i]$, we begin by marginalizing out all independent variables from \eqref{eq:conj_post_proof}, resulting in:

\begin{align}
    \E[\mu_i] &= \int_{\mu_i} \frac{\mu_i}{M} \Big[ c_i \NIG(\mu_i, \sigma^2_i | \Tilde{\tau}_i, \Tilde{\kappa}_i, \Tilde{\beta}_i, \Tilde{\gamma}_i) + \nonumber \\
    & \;\;\;\;\;+ \sum_{j \neq i}^k c_j \NIG(\mu_i, \sigma^2_i | \tau_i, \kappa_i, \beta_i, \gamma_i) \Big] \mathrm{d}\mu_i \\
    &= \frac{1}{M} \Big[ c_i \int_{\mu_i} \mu_i \NIG(\mu_i, \sigma^2_i | \Tilde{\tau}_i, \Tilde{\kappa}_i, \Tilde{\beta}_i, \Tilde{\gamma}_i) \mathrm{d}\mu_i + \nonumber \\
    & \;\;\;\;\;+ \sum_{j \neq i}^k c_j \int_{\mu_i} \mu_i \NIG(\mu_i, \sigma^2_i | \tau_i, \kappa_i, \beta_i, \gamma_i) \mathrm{d}\mu_i\Big] \\
    &= \frac{1}{M} \Big[ c_i \Tilde{\tau_i} + \sum_{j \neq i}^k c_j \tau_i \Big]
\end{align}

where the first equality follows from the linearity of the expectation operator and the second equality follows from the the fact that for a Normal-Inverse-Gamma distribution, $\NIG(\mu,\sigma^2|\tau,\kappa,\beta,\gamma)$, the expected value for $\mu$ is simply $\E[\mu] = \tau$.

\section{Additional Implementation Details} \label{app:C}

This section describes the calibration of the uncertainty of the semantic segmentation network and the class-wise Gaussian model used to probabilistically relate material class to friction.

The pre-trained SegFormer+FastSAM network discussed in Section \ref{sec:implementation} is evaluated on the same test set from the Dense Material Segmentation Dataset \cite{upchurch2022materials} used in the simulation experiments in Section \ref{subsec:dmsd_eval}.
The confusion matrix for this evaluation is shown in Figure \ref{fig:confusion_matrix}.
From this confusion matrix we note the semantic segmentation network struggles to correctly predict plastic, concrete, and grass, however it is able to accurately predict the other material classes with high likelihood.
Notably, the biggest source of error in the network is from incorrectly predicting the \textit{None} class rather than classification as a different material.

Only a subset of the Dense Material Segmentation Dataset classes are considered in this work.
The remaining classes are clustered into the \textit{None} category, resulting in a class imbalance in the dataset.
This is the most likely cause for the incorrect \textit{None} class predictions.

With respect to the Gaussian property model calibration, we refer to \cite[\S VII]{ewen2022these} which describes how this class-wise Gaussian model was chosen and calibrated.

\end{document}